\PassOptionsToPackage{table,xcdraw}{xcolor}
\documentclass[11pt]{article}

\usepackage[final]{acl}

\usepackage{times}
\usepackage{latexsym}
\usepackage[T1]{fontenc}
\usepackage[utf8]{inputenc}
\usepackage{microtype}
\usepackage{inconsolata}
\usepackage{graphicx}

\usepackage{setspace}
\usepackage{listings}
\usepackage[ruled]{algorithm}
\usepackage{algpseudocode}
\usepackage{amsmath}
\usepackage{amssymb}
\usepackage{booktabs}
\usepackage{multirow}
\usepackage{float}
\usepackage{colortbl,pgf}
\usepackage{newfloat}
\usepackage{tikz}
\usepackage{subfigure}
\usepackage{caption}
\usepackage{enumitem}
\usepackage[most]{tcolorbox}
\tcbuselibrary{breakable}
\usepackage{xfp}

\newtcolorbox{responsebox}[1]{
  enhanced,
  breakable,
  colback=green!5!white,
  colframe=green!55!black,
  coltitle=white,
  fonttitle=\bfseries,
  title=#1,
  arc=2mm,
  boxrule=1pt,
  width=\textwidth
}

\newtcolorbox{promptbox}[1]{
  enhanced,
  breakable,
  colback=blue!5!white,
  colframe=blue!75!black,
  coltitle=white,
  fonttitle=\bfseries,
  title=#1,
  arc=2mm,
  boxrule=1pt,
  width=\textwidth
}

\newcommand{\framework}{\textsc{EvoSQL}}

\newcommand{\scorecell}[1]{%
  \begingroup
  \edef\colval{\fpeval{max(0, min(100, round(#1-50)))}}%
  \ifnum\colval>0\relax
    \edef\doCell{\noexpand\cellcolor{blue!\colval}}%
    \doCell #1%
  \else
    \cellcolor{white}#1%
  \fi
  \endgroup
}

\newcommand{\percentcell}[1]{%
  \begingroup
  \ifdim #1pt>0pt\relax
    \colorbox{green!15}{\scriptsize $#1$}%
  \else\ifdim #1pt<0pt\relax
    \colorbox{red!15}{\scriptsize $#1$}%
  \else
    \colorbox{gray!10}{\scriptsize $#1$}%
  \fi\fi
  \endgroup
}

\title{\framework{}: Memory-Augmented Critic-Generator Co-Evolution for Text-to-SQL}

\author{
  Jiawei Zhou \textsuperscript{1,2},
  Jianwei Wang \textsuperscript{4}\thanks{Corresponding Authors. Email: \url{jianwei.wang1@unsw.edu.au} (Jianwei Wang) and \url{w.kai@sjtu.edu.cn} (Kai Wang)} 
  Chenyu Zhou \textsuperscript{3}, Chaojian Shi \textsuperscript{1}, Ming Dong \textsuperscript{5}, 
  Kai Wang \textsuperscript{3}\footnotemark[1] \\
  \\
  \textsuperscript{1} Fudan University 
  \textsuperscript{2} Shanghai AI Lab 
  \textsuperscript{3} Shanghai Jiao Tong University \\
  \textsuperscript{4} University of New South Wales 
  \textsuperscript{5} Wuhan University of Technology 
} 

\begin{document}
\maketitle

\begin{abstract}
Text-to-SQL has advanced rapidly with large language models, but complex database queries still require reasoning beyond one-shot generation, including multi-step decomposition, execution-based diagnosis, and targeted correction. We present \framework{}, a co-evolution framework that formulates SQL synthesis as an iterative interaction between a generator and a critic.
\framework{} maintains a contextualized candidate memory, verifies SQL candidates with both execution signals and LLM-based critique, and updates its memory through utility-guided aggregation. To strengthen the underlying generator–critic pair, we further introduce a Self-Distillation Policy Optimization (SDPO) fine-tuning stage that injects execution-aware supervision into modern coding LLM backbones. Experiments on Spider and BIRD show that \framework{} consistently improves open-source models over Maj@16 baselines, with particularly large gains on BIRD-Dev, ranging from +1.37\% for Qwen3-4B to +9.19\% for Qwen2.5-Coder-3B. SDPO initialization further improves selected backbones on Spider-Test and BIRD-Dev. These results suggest that memory-grounded co-evolution is an effective path toward more reliable and generalizable Text-to-SQL systems.
Code is available at \url{https://github.com/valleysprings/EvoSQL}.
\end{abstract}

\section{Introduction}
\label{main_introduction}

Text-to-SQL (Text2SQL), also known as Natural Language to SQL, enables users to query relational databases using natural language, substantially lowering the barrier to data access and analysis~\citep{hong2024next}. With the rise of large language models (LLMs), modern Text2SQL systems have rapidly advanced along multiple directions, including schema-aware modeling and schema linking~\citep{resdsql, cao2024rsl}, content-aware retrieval and schema selection~\citep{chess}, and test-time correction or multi-step prompting pipelines~\citep{dinsql}. 

\begin{figure}[ht]
    \centering
    \includegraphics[width=\linewidth]{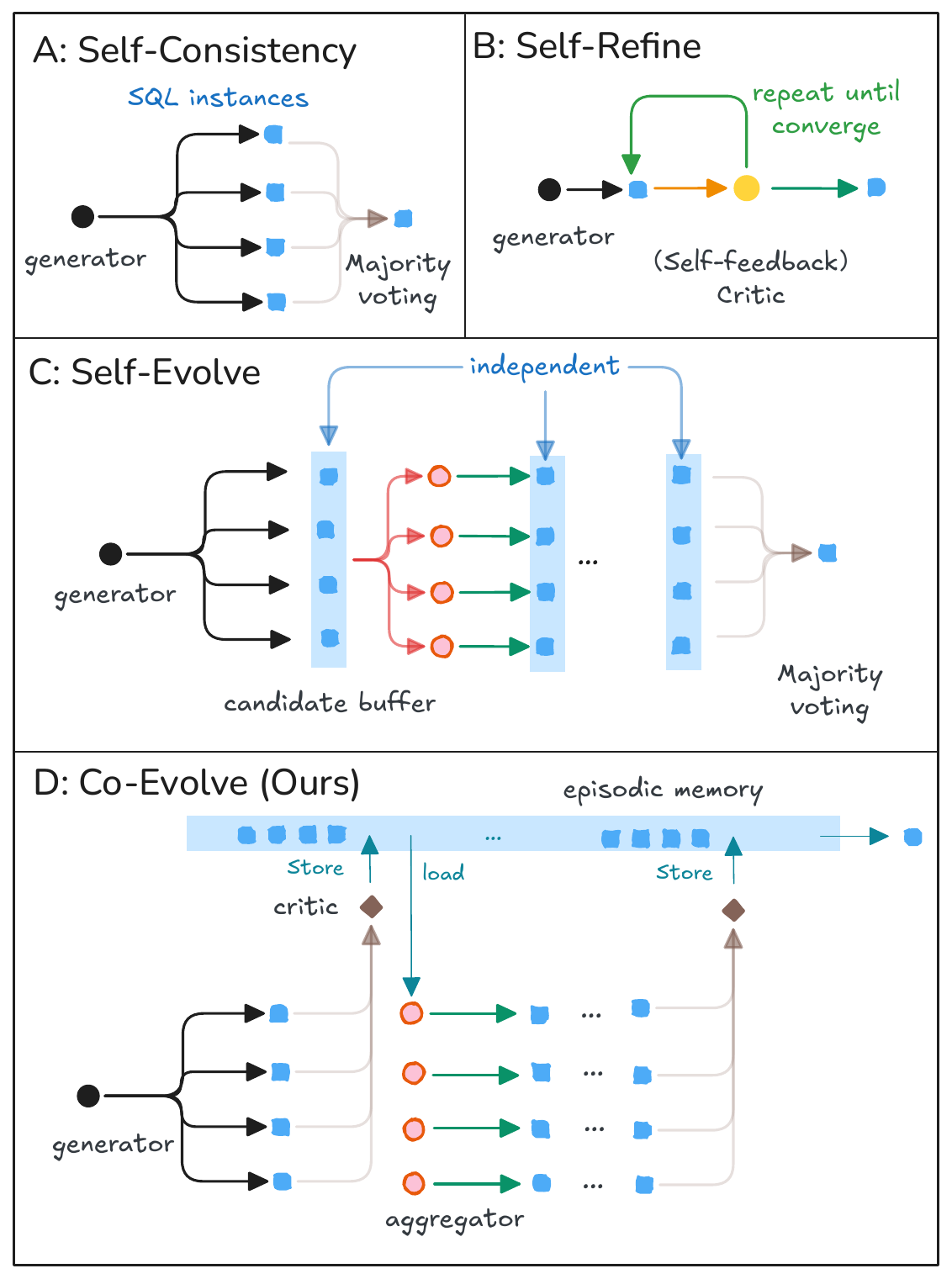}
    \caption{\textbf{Paradigms for Text2SQL.}  (A) Self-Consistency: sample many candidates and select by majority/critic. (B) Self-Refine: Iteratively critique and revise a single candidate. (C) Self-Evolve: Aggregates SQL candidates across generations, then performs majority voting. (D) Co-Evolve (Ours): Multi-round population evolution with contextualized episodic memory (store/load), where a critic provides reusable diagnoses and an aggregator performs targeted updates, followed by greedy selection.}
    \label{fig:example}
\end{figure}

Despite these successes, existing Text2SQL systems still face significant bottlenecks when reasoning over complex database scenarios (e.g., multi-hop reasoning, nested queries). Current SOTA methods can be roughly divided into two families: prompt-based inference-time methods and learning-based optimization methods.

\paragraph{Prompt-based methods}
Prompt-based methods improve off-the-shelf LLMs mainly at inference time without explicitly updating the underlying model. 
Single-agent pipelines typically rely on in-context generation, self-consistency, or self-refinement, where one model samples, critiques, and revises SQL candidates through prompting~\citep{dailsql, c3sql, dinsql}. 
Multi-agent pipelines further decompose Text2SQL into specialized roles such as planning, schema exploration, generation, and verification, or allocate test-time compute across multiple reasoning paths before selecting by agreement or preference (e.g., MAC-SQL, ReFoRCE, CHASE-SQL)~\citep{macsql, deng2025reforce, chasesql}. 
These methods increase structural diversity and enable iterative correction, but their search process is usually governed by fixed prompt templates, fixed candidate budgets, or heuristic role interactions. 
This creates two challenges on complex schemas: first, they lack difficulty-aware control, wasting compute on easy cases while providing insufficient exploration for hard ones; second, without effective contextualized memory, they often forget previous failure modes, revisit the same erroneous conditions, or oscillate between superficially plausible SQL variants. 
As a result, prompt-based multi-agent methods may still drift or spend substantial budget on unfocused corrections in complex scenarios.

\paragraph{Learning-based methods}
Learning-based methods instead improve Text2SQL models by updating their parameters or policies. 
Most existing Text2SQL optimization is still \emph{single-turn}: supervised fine-tuning learns from curated or synthesized SQL data, such as OmniSQL~\citep{omnisql}; iterative self-training distills pseudo-labeled rationales or high-confidence solutions back into the model, such as STaR-SQL~\citep{he2025star}; and RL with verifiable rewards (RLVR) directly optimizes execution-grounded correctness, with SQL-R1 as a representative~\citep{ma2025sql}. 
While effective, these methods are optimized to produce a complete SQL query in one generation step, with only limited post-hoc aggregation such as self-consistency. 
For hard queries, however, correctness often requires localized semantic guidance, trial-and-error execution feedback, and critique-driven revision to determine whether a candidate condition is valid. 
Moreover, RLVR may further induce \emph{diversity collapse}~\citep{chen2026does}, where the policy converges to a narrow family of SQL structures that are locally optimal under the reward, reducing exploration of alternative but valid reasoning routes. 
Beyond single-turn learning, emerging multi-turn optimization paradigms such as recursive self-aggregation aim to evolve a population by aggregating candidates across generations~\citep{venkatraman2025recursiveselfaggregationunlocksdeep}. 
However, multi-turn optimization remains largely unexplored for Text2SQL and is inherently challenging, owing to the unique complexity of the task, such as tight schema grounding, execution feedback, and structured candidate refinement. How to leverage multi-turn experience for informed and continuously improving SQL generation remains an open challenge.

\begin{figure*}[ht]
    \centering
    \includegraphics[width=\linewidth]{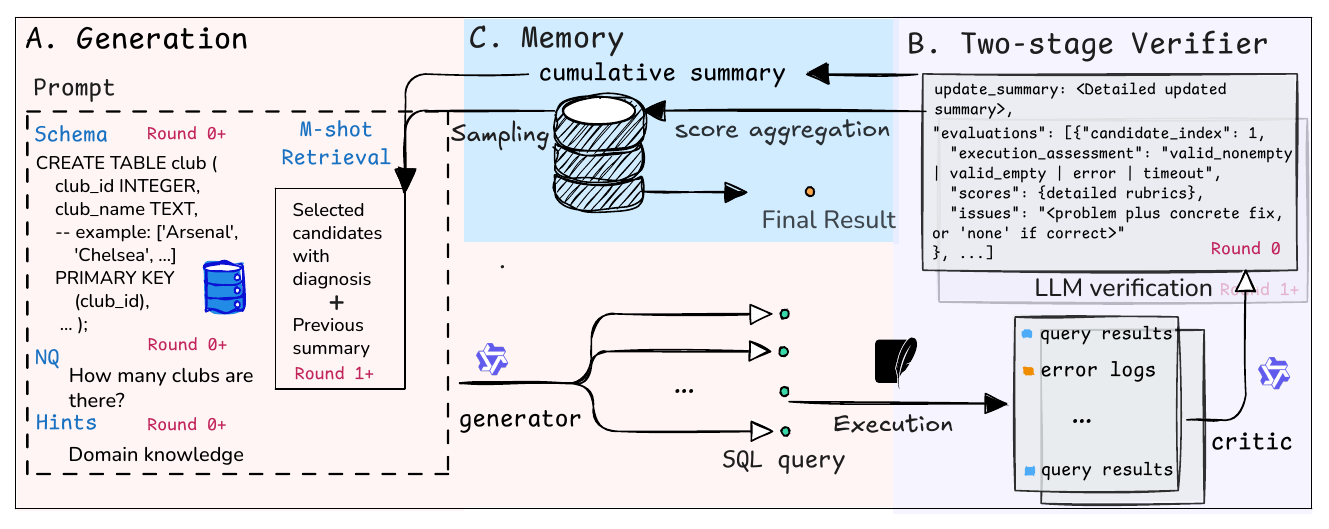}
    \caption{\textbf{Online inference-time workflow of \framework{}.} 
    (A) The generator elicits parallel SQL candidates from the question, schema, hints and M-shot candidates from memory (Round 1+).
    (B) A two-stage verifier combines deterministic execution feedback with LLM-based critique to score and diagnose candidates.
    (C) The resulting memory stores verified candidates and summaries, which are
    used for utility-guided payload sampling, aggregation, and final SQL selection.
    }
    \label{fig:pipeline}
\end{figure*}

To address this gap, we propose \framework, a novel Co-Evolving SQL Generation Framework, as illustrated in Figure~\ref{fig:pipeline}. Unlike prompt-only orchestration methods~\citep{macsql,deng2025reforce,chasesql} or learning-based single-turn optimization methods~\citep{omnisql,he2025star,ma2025sql}, our approach couples a generator and a critic in a multi-round evolutionary process, where SQL candidates first undergo dual-stage verification, are then utility-ranked for selection, and are finally improved through aggregation over selected candidates with contextualized memory.

Specifically, we design a Co-Evolution mechanism in which SQL candidates are first verified through execution signals and critic feedback, then selected according to their utility, and finally aggregated into improved candidates with support from an episodic memory of scored experiences. To ensure both diversity and search stability, we integrate execution information into the selection process, guiding evolution toward logically consistent and executable solutions. For hard samples where the LLM critic may not provide sufficient feedback to effectively guide exploration,
we also study an induction-oriented Self-Distillation Policy Optimization (SDPO) for strengthening SQL synthesis before inference-time Co-Evolution.

Main contributions are summarized as follows:

\begin{itemize}
\item We propose \framework{}, a Co-Evolving Text-to-SQL framework that introduces memory-augmented multi-round candidate evolution for SQL generation.
\item \framework{} combines execution feedback, critic-guided refinement, utility-based selection, episodic memory, and early stopping to efficiently explore and aggregate diverse SQL candidates.
\item We study an induction-oriented SDPO stage that distills execution-aware preference signals into post-trained backbones, improving SQL synthesis and revision on challenging queries.
\item Experiments on Spider and BIRD demonstrate consistent gains over Maj@16 baselines, while ablations and sensitivity analyses confirm the effectiveness and robustness of the proposed Co-Evolution mechanism.
\end{itemize}

\section{Related Work}
\label{main_relatedwork}

\paragraph{Current Text-to-SQL paradigms}
Recent advances in large language models (LLMs) have substantially propelled
Text-to-SQL (or NL2SQL) systems~\citep{katsogiannis2023survey, hong2024next}.
Existing methods typically improve different stages of the Text-to-SQL pipeline.
Schema-aware methods focus on schema linking, schema pruning, and structural
encoding to better align user questions with tables and columns~\citep{resdsql};
decomposition-based methods break complex questions into simpler sub-problems
or intermediate representations~\citep{scprompt, zeronl2sql, natsql,
wolfson2020break, eyal2023semantic, gao2022towards, petsql}; and
post-processing methods use execution feedback, self-correction, or
consistency-based selection to repair or select SQL candidates~\citep{liu2025nl2sql,
dinsql, macsql, dailsql, c3sql, omnisql}.

Learning-based Text-to-SQL systems further adapt LLMs through supervised
fine-tuning, synthetic data construction, or reinforcement learning. SFT-based
approaches can improve in-domain accuracy, but they often depend on static
training distributions and may generalize poorly to unseen schemas or
database-specific reasoning patterns~\citep{dtssql, catsql, ratsql, spider2}.
More recently, RLVR-style methods such as SQL-R1~\citep{ma2025sql} use execution
correctness as a verifiable reward, enabling SQL models to acquire stronger
reasoning and error-recovery behaviors than standard SFT. However, most existing
learning-based methods still optimize a single-turn generator: the model is
trained to produce a complete SQL query in one pass, and inference-time
improvement is usually limited to resampling, voting, or shallow correction.
As a result, they do not explicitly maintain reusable candidate-level
diagnoses, nor do they couple a generator and a critic in a multi-round
Co-Evolution process.

\paragraph{Multi-turn Self-Evolution and Self-Distillation for code agents.}
Beyond Text-to-SQL, recent work on code-centric reasoning has shown that
executable environments can serve as effective feedback sources for improving
LLM agents. In Self-Evolution and some Self-Play paradigms, models iteratively
propose solutions, interact with an environment, receive outcome-level feedback,
and improve their future behavior through reinforcement learning or
trajectory-level selection~\citep{deepseekr1, huang2025r, zhao2025absolute}.
This line of work is especially effective for code tasks because compilers,
interpreters, unit tests, and execution traces provide objective signals for
both correctness checking and error diagnosis.

A related trend studies multi-turn interaction as a mechanism for building more
robust agentic systems. Self-Evolution and recursive aggregation methods
maintain populations of candidate solutions and improve them across rounds via
selection, recombination, or self-aggregation~\citep{novikov2025alphaevolve,
lange2025shinkaevolve, venkatraman2025recursiveselfaggregationunlocksdeep}.
Tool-integrated RL further trains agents to discover stable interaction
protocols with external tools or environments, improving long-horizon reasoning
and tool use~\citep{deepretrieval, searchr1, li2025torl, wang2025ragen}.
Meanwhile, self-distillation policy optimization methods~\citep{hubotter2026reinforcement, zhao2026self} and GRPO-based variants~\citep{li2026unifying} aim to transfer
stronger trajectories, verifier feedback, or teacher preferences back into the
policy, thereby improving the model's ability to plan, revise, and recover from
errors across interaction rounds.

Together, these works suggest that executable feedback and multi-turn
interaction can substantially improve code-oriented reasoning. Building on this
insight, \framework{} instantiates Co-Evolution for Text-to-SQL by combining
generator--critic interaction, execution feedback, candidate memory, and an optional SDPO alignment.

\section{Problem Formulation}
\label{sec:formulation}

Given a natural-language question $x$, its associated gold SQL query $q$, a database schema $\mathcal{S}=(\mathcal{T},\mathcal{C},\mathcal{R})$ containing tables, columns, and foreign-key relations, and a database instance $\mathcal{D}$, the goal of NL2SQL is to generate an executable SQL query $\hat{q}$ whose execution outcome matches that of the ground truth $q$. Formally, a prediction is correct if
\begin{align}
\operatorname{Exec}(\hat{q} \mid\mathcal{D},\mathcal{S})
\equiv
\operatorname{Exec}(q \mid\mathcal{D},\mathcal{S})
\end{align}
and the overall objective is to maximize execution correctness over the evaluation set.

To improve inference-time performance, we consider a multi-round test-time scaling procedure. At round $t$, a generator $\mathcal{M}_\theta$ produces a candidate SQL query
\begin{align}
q_{(i,t)} \sim \mathcal{M}_\theta(\cdot \mid x, \mathcal{D},\mathcal{S},k_{(i,t)}),
\end{align}
where $k_{(i,t)}$ denotes the available mutation context (or few-shot payloads) at round $t$. For the initial round, $k_{(i,1)}$ is empty; for later rounds, $k_{(i,t)}$ may additionally include information derived from previously generated candidates.

Each candidate is then processed by an execution module, which produces a deterministic execution record
\begin{align}
e^d_{(i,t)}:= \operatorname{Exec}(q_{(i,t)}|\mathcal{D},\mathcal{S}),
\end{align}
where $e^d_{(i,t)}$ may include the execution result, execution status, and runtime error messages.

Second, an LLM-based critic $\mathcal{M}_\phi$ produces a structured critique conditioned on the question, schema, candidate query, and execution record:
\begin{align}
e^p_{(i,t)} \sim \mathcal{M}_\phi(\cdot \mid x,\mathcal{D},\mathcal{S},q_{(i,t)},e^d_{(i,t)}).
\end{align}

This critique includes a rubric score vector (e.g., syntax, schema consistency, logical correctness, and completeness) and optional mutation instructions.

\section{Methodology}
\label{sec:methodology}

\subsection{Framework Overview}

We propose \textbf{\framework} (Co-\textbf{Evo}lution-based Text2\textbf{SQL}), a Text2SQL-oriented test-time harness that wraps a base SQL generator with execution, critique, memory, and selection modules. Rather than replacing existing Text2SQL models, \framework{} is designed to be model-agnostic: a single-turn optimized model, such as an SFT- or RLVR-trained generator, can be directly plugged into the harness and further evolved at inference time. The core online process is shown in Figure~\ref{fig:pipeline}; an optional offline SDPO alignment stage can further improve the generator with execution-aware aggregation feedback.

Compared with execution-only Self-Evolution, which mainly reuses previously generated candidates and database feedback, \framework{} introduces an explicit critic--generator Co-Evolution loop. The critic does not generate SQL; instead, it diagnoses candidate programs based on deterministic execution. These diagnoses are converted into calibrated utilities and reusable memory, allowing the generator to perform targeted revision rather than unconstrained resampling. This additional critic role is the key distinction from Self-Evolution: it provides clause-level guidance, filters misleading executable queries, and helps decide when further evolution is unnecessary.

The online workflow follows three steps. First, \framework{} performs an elicitation step, where the generator samples multiple initial SQL candidates for the given question and schema. Second, each candidate is executed and then evaluated by the critic; the resulting execution records, scores, and a round summary are stored in the candidate memory. Third, in later rounds, \framework{} retrieves high-utility and informative candidates from memory as few-shot payloads, asks the generator to synthesize improved SQL candidates, and repeats verification, memory update, and utility-based selection until early stopping or the round budget is reached. Finally, the best candidate based on the utility score is selected from the accumulated memory.

\subsection{Multi-Round Episodic Memory}
\label{sec:replay_memory}

For each problem, \framework{} maintains a memory pool $\mathcal{Q}=\{\hat{q}_{(i,t)} \mid i\in[K],\ t\in[T]\}$, where $K$ denotes the sampling number per round and $T$ denotes maximal evolving rounds.

\subsubsection{Utility function for Payload Selection}

\paragraph{Execution-grounded verifier as candidate priority}
Execution feedback is precise for catastrophic failures (syntax error, schema error, timeout, empty set) but insufficient to rank semantically close candidates that all execute.
We therefore use an LLM-based critic/verifier $\mathcal{M}_\phi$ that parses $c^t_i$ and emits either a single scalar score or a vector of rubric scores. Let $\mathbf{s}_{(i,t)}$ denote the returned scores and $m=|\mathbf{s}_{(i,t)}|$ for this candidate. We first compute a raw score by averaging multiple rubric dimensions, or by directly using the single score when only one dimension is returned:
\begin{align}
s_{\text{raw}}(q_{(i,t)})=
\begin{cases}
\frac{1}{m}\sum_{\ell=1}^{m} s_{(i,t),\ell}, & m>1,\\
s_{(i,t),1}, & m=1.
\end{cases}
\end{align}

\noindent To prevent ungrounded over-confidence, we calibrate with execution-based caps:
\begin{align}
\text{Conf}(q_{(i,t)})=
\begin{cases}
-1, & e^d_{(i,t)} \in \mathcal{E}_{\text{invalid}}\\
0,  & e^d_{(i,t)} \in \mathcal{E}_{\text{none}}\\
s_{\text{raw}}(q_{(i,t)}), & \text{otherwise}
\end{cases}
\end{align}
Specifically, invalid SQL is suppressed and should be refined only if all candidates are invalid, and empty outputs are explicitly down-weighted.

Multi-round refinement must balance exploration (trying diverse structures, such as alternative joins and nesting) and exploitation (refining promising hypotheses).
\framework{} instantiates this balance through a time-aware utility and a consistency bonus.

\paragraph{Time-discounted utility and consistency bonus}
For each current timestamp $t$, we rank all candidates stored in $\mathcal{Q}_{\le t}$ based on the given utility (Fig.~\ref{fig:pipeline}). Let $q_{(i,t')}$ denote a candidate whose latest occurrence timestamp is $t'\le t$. If the same SQL string appears again in a later round, we update $t'$ to its latest occurrence rather than treating the older copy as a separate fresh candidate. The utility is defined as
\begin{align}
U_t(q_{(i,t')}) &= \gamma^{(t - t')}\cdot \mathrm{Conf}(q_{(i,t')}) \\
&+ \lambda_{\mathrm{cons}}\cdot \mathrm{Cons}_t(q_{(i,t')}),
\label{eq:utility_timestamp}
\end{align}
where $\gamma\in(0,1]$ down-weights stale candidates while preserving alternative paths, and $\mathrm{Cons}_t(q_{(i,t')})$ functions as a majority-based bonus. Specifically,
\begin{align*}
\mathrm{Cons}_t(q_{(i,t')})
= \log\big(1 + N_t(q_{(i,t')})\big),
\end{align*}
where $N_t(q_{(i,t')})$ is the number of candidates in $\mathcal{Q}_{\le t}$ that share the same execution result as $q_{(i,t')}$ and have confidence of at least $\tau$. This term increases when multiple high-confidence candidates agree on the same execution outcome.

We then sample candidates from the append-only pool $\mathcal{Q}_{\le t}$ according to their utility:
\begin{align}
\mathbb{P}(q_i \mid \mathcal{Q})
=
\frac{\exp\!\left(U(q_i)/\kappa\right)}
{\sum_{q_j \in \mathcal{Q}} \exp\!\left(U(q_j)/\kappa\right)},
\end{align}
where $\kappa > 0$ controls the exploration--exploitation trade-off. We sample $M$ candidates as payload $k_{(i,t)}$.

\subsubsection{Early Stopping}

We stop refinement once the candidate pool exhibits both high confidence and stable execution outcomes. Let $\mathcal{B}_t=\operatorname{TopK}(\mathcal{Q})$. Concretely, at timestamp $t$, we terminate early if there exists an execution outcome $r^*$ such that
\begin{align}
\frac{
\left|
\left\{
q \in \mathcal{B}_t :
d(q)=r^*,\ \mathrm{Conf}(q)\ge \tau
\right\}
\right|
}{
\left|\mathcal{B}_t\right|
}
\ge \pi^\star,
\end{align}
where $\pi^\star $ is the stopping ratio.
To produce the final SQL, we select the execution cluster with the top utility score from $\mathcal{Q}$.

\subsection{Co-Evolution Fine-tuning with Rich Feedback}
\label{sec:coevolution}

Our Co-Evolution framework couples a Generator $\mathcal{M}_\theta$ that proposes
SQL candidates with a Critic $\mathcal{M}_\phi$ that diagnoses semantic and
execution-level errors. Here, we apply an offline SQL-oriented SDPO stage mainly to
strengthen SQL synthesis. Using SQL-R1 data~\citep{ma2025sql}, we construct
privileged teacher signals from the gold SQL $q$ and its execution result
$e^d=\operatorname{Exec}(q\mid\mathcal{D},\mathcal{S})$, and distill this
execution-aware supervision into the model.

\subsubsection{Objective Alignment}

The offline alignment focuses on the induction side, i.e., evolving the generator on SQL synthesis. For each SQL-R1 training instance, we take the natural-language question, schema, database, gold query, and the gold query's execution result as the supervision tuple. The generator is trained to predict SQL tokens directly under the original problem context, while the teacher additionally observes the correct query and its execution output.

\paragraph{SQL-Induction Construction}
For a training SQL trajectory $\hat{q}$ and each SQL-token prefix $\hat{q}_{<\ell}$, we define the student and teacher next-token distributions as
\begin{align}
P_\ell^S &=\mathcal{M}_\theta^S(\cdot \mid x,\mathcal{D},\mathcal{S},\hat{q}_{<\ell}) \\
P_\ell^T &=\mathcal{M}_\theta^T(\cdot \mid x,\mathcal{D},\mathcal{S},\hat{q}_{<\ell},q,e^d)
\end{align}
where $q$ and $e^d=\operatorname{Exec}(q\mid\mathcal{D},\mathcal{S})$ are privileged signals available only to the teacher during offline alignment. We then apply logit-level on-policy distillation:
\begin{align}
\mathcal{L}_{\mathrm{SDPO}}^{\mathrm{SQL}}
&:= \mathbb{E}_{\mathcal{T}}
\Bigg[
\frac{1}{|\hat{q}|}\sum_{\ell=1}^{|\hat{q}|}
\mathrm{KL}\big(P_\ell^S\,\|\,\operatorname{sg}(P_\ell^T)\big)
\Bigg],
\label{eq:sql_opd}
\end{align}
where $\mathcal{T}$ denotes the offline alignment data and $\operatorname{sg}(\cdot)$ denotes the stop-gradient operator. Thus, SDPO transfers the teacher's execution-aware preference into next-token SQL prediction.

\section{Experiments}
\label{main_exp}

\subsection{Setup}
\label{setup}

\paragraph{Evaluation Benchmarks}
We evaluate \framework{}  on two widely used NL2SQL benchmarks: Spider~\citep{spider} (dev/test) and BIRD~\citep{bird} (dev only). Spider contains 10,181 questions paired with 5,693 complex SQL queries over 200 databases, while BIRD consists of 12,751 NL2SQL pairs across 95 databases.

\paragraph{Metric}
We report \emph{Execution Accuracy} (EX). Prediction $\hat{q}$ is counted as correct if it is executable and its denotation matches the gold query $q$. EX is computed as the number of correct predictions divided by the total number of examples. We enforce a 30-second timeout with single-core execution for each query on both Spider and BIRD.

For all methods, we use an identical schema serialization to ensure input integrity and fair comparison. The input sequence concatenates the question and the database schema formatted as \texttt{CREATE TABLE} statements, augmented with column attribute descriptions and representative values following prior work~\citep{chess, omnisql, yang2024synthesizing, rajkumar2022evaluating}.

\paragraph{Compared Settings}
Our primary focus is to compare \emph{evolution at inference time}. We therefore establish strong model-specific baselines and then measure gains brought by \framework.
\textbf{(i) Maj@16} denotes the backbone model's best-effort decoding baseline using \emph{self-consistency} with majority voting where $K{=}16$.
\textbf{(ii) \framework{}} (Base) runs our full test-time evolution framework on the same backbone, using the same schema input format and candidate budget.
\textbf{(iii) \framework{} (SDPO)} further applies SDPO fine-tuning to selected base backbones before running \framework.

\paragraph{Implementation Details}
We evaluate four backbone settings: Coder-3B, Coder-7B, Qwen3-4B, and SQL-R1~\citep{ma2025sql}. For inference, we sample $16$ SQL candidates with temperature $0.8$ and keep at most $3$ evolution rounds, since additional rounds yield diminishing marginal gains.

In \framework{} (SDPO), we only fine-tune two base backbones, Coder-3B and Qwen3-4B. We use a two-stage SDPO procedure: the first stage extracts execution feedback $e^d$, and the second stage applies SDPO to refine the model. We train on a 5k-example selection set derived from the SQL-R1 SynSQL-5k set~\citep{ma2025sql}.

\paragraph{Environment}
All experiments are conducted on Ubuntu 22.04 LTS with 128 Intel CPU cores, 1 TB of system memory, and 4 NVIDIA H200 GPUs.

\subsection{Main Results}

\subsubsection{Performance on Main Benchmarks}

Table~\ref{tab:mainresult} reports the strict execution accuracy (EX) on Spider-Dev, Spider-Test, and BIRD-Dev. Under the same Maj@16 sampling budget, \framework{} (Base) consistently improves all four backbones across all three benchmarks, indicating that the gain is not tied to a specific model family.
The improvement is especially clear on BIRD-Dev, where \framework{} obtains
+9.19 EX for Coder-3B and steady gains for stronger backbones, including
+1.44 for Coder-7B, +1.37 for Qwen3-4B, and +1.49 for SQL-R1. This suggests
that critic-guided Co-Evolution is particularly useful for harder
database-grounded queries where majority voting alone is insufficient.

\begin{table}[ht]
    \centering
    \caption{Strict execution accuracy~(\%) on Spider and BIRD benchmarks. 
    For both \framework{} (Base) and \framework{} (SDPO), $\Delta$ denotes the
percentage-point difference relative to the corresponding Maj@16 result.}
    \label{tab:mainresult}
    \resizebox{\columnwidth}{!}{%
    \begin{tabular}{@{}ccccc@{}}
    \toprule
        \textbf{Setting} & \textbf{Backbone} & \textbf{Spider-Dev} & \textbf{Spider-Test} & \textbf{BIRD-Dev} \\
    \midrule
    \multirow{4}{*}{\textbf{Maj@16}}
          & Coder-3B & 76.40 & 76.01 & 51.24 \\
          & Coder-7B & 80.08 & 81.23 & 61.60 \\
          & Qwen3-4B & 84.53 & 82.35 & 65.19 \\
          & SQL-R1 & 83.08 & 83.74 & 65.65 \\
    \midrule
    \multirow{8}{*}{\shortstack{\textbf{\framework}\\(Base)}}
          & Coder-3B & 77.85 & 76.99 & 60.43 \\
          & $\Delta$ & \percentcell{+1.45} & \percentcell{+0.98} & \percentcell{+9.19} \\
          & Coder-7B & 81.53 & 81.32 & 63.04 \\
          & $\Delta$ & \percentcell{+1.45} & \percentcell{+0.09} & \percentcell{+1.44} \\
          & Qwen3-4B & 84.82 & 82.95 & 66.56 \\
          & $\Delta$ & \percentcell{+0.29} & \percentcell{+0.60} & \percentcell{+1.37} \\
          & SQL-R1 & 83.46 & 84.63 & 67.14 \\
          & $\Delta$ & \percentcell{+0.38} & \percentcell{+0.89} & \percentcell{+1.49} \\
    \midrule
    \multirow{4}{*}{\shortstack{\textbf{\framework}\\(SDPO)}}
          & Coder-3B & 76.02 & 79.41 & 60.89 \\
          & $\Delta$ & \percentcell{-0.38} & \percentcell{+3.40} & \percentcell{+9.65} \\
          & Qwen3-4B & 82.88 & 84.07 & 67.08 \\
          & $\Delta$ & \percentcell{-1.65} & \percentcell{+1.72} & \percentcell{+1.89} \\
    \bottomrule
    \end{tabular}
    }
\end{table}

With SDPO initialization, \framework{} further improves Spider-Test and
BIRD-Dev for the two fine-tuned backbones: +3.40/+9.65 EX for Coder-3B and +1.72/+1.89 EX for Qwen3-4B. The slight drop on Spider-Dev suggests that SDPO may bias the model toward
harder execution-aware correction patterns, which do not uniformly benefit easier queries.
Additional results are reported in Appendix~\ref{app:full_sql_bench}.

\subsubsection{Performance on Multi-Turn Co-Evolution}

Table~\ref{tab:round_improvement_strict} compares the first-round Co-Evolved prediction with the final Co-Evolved prediction. Overall, the final prediction improves over T1 by $+1.37$ points on average, with gains of $+2.56$, $+0.39$, and $+1.17$ points on Spider-Dev, Spider-Test, and BIRD-Dev, respectively.

\begin{table}[ht]
      \centering
      \caption{Round-wise strict execution accuracy~(\%) of multi-turn Co-Evolution. T1 denotes the first-round
  Co-Evolved prediction, and T3 denotes the final Co-Evolved prediction.}
      \label{tab:round_improvement_strict}
      \resizebox{\columnwidth}{!}{%
      \begin{tabular}{@{}ccccc@{}}
      \toprule
          \textbf{Backbone} & \textbf{Round} & \textbf{Spider-Dev} & \textbf{Spider-Test} & \textbf{BIRD-Dev} \\
      \midrule
          \multirow{3}{*}{Coder-3B} & T1 & 74.47 & 76.71 & 55.28 \\
              & T3 & 77.85 & 76.99 & 60.43 \\
              & $\Delta$ & \percentcell{+3.38} & \percentcell{+0.28} & \percentcell{+5.15} \\
          \midrule
          \multirow{3}{*}{Coder-7B} & T1 & 78.82 & 80.53 & 63.75 \\
              & T3 & 81.53 & 81.32 & 63.04 \\
              & $\Delta$ & \percentcell{+2.71} & \percentcell{+0.79} & \percentcell{-0.71} \\
          \midrule
          \multirow{3}{*}{Qwen3-4B} & T1 & 82.11 & 83.37 & 66.82 \\
              & T3 & 84.82 & 82.95 & 66.56 \\
              & $\Delta$ & \percentcell{+2.71} & \percentcell{-0.42} & \percentcell{-0.26} \\
          \midrule
          \multirow{3}{*}{SQL-R1} & T1 & 82.01 & 83.70 & 66.62 \\
              & T3 & 83.46 & 84.63 & 67.14 \\
              & $\Delta$ & \percentcell{+1.45} & \percentcell{+0.93} & \percentcell{+0.52} \\
      \midrule
          \multirow{3}{*}{\textbf{Average}} & T1 & 79.35 & 81.08 & 63.12 \\
              & T3 & 81.92 & 81.47 & 64.29 \\
              & $\Delta$ & \percentcell{+2.56} & \percentcell{+0.39} & \percentcell{+1.17} \\
      \bottomrule
      \end{tabular}
      }
  \end{table}

\subsection{Sensitivity Analysis}

\begin{figure}[h]
    \centering
    \includegraphics[width=\linewidth]{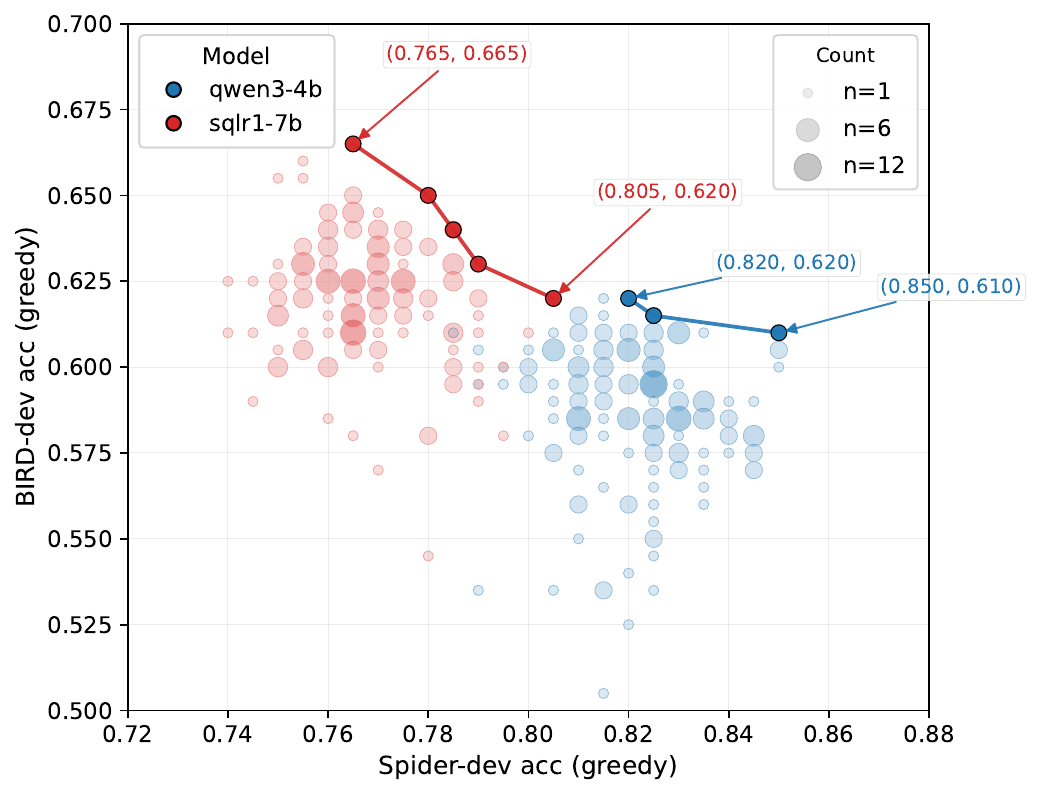}
    \caption{\textbf{Sensitivity analysis of Co-Evolution hyperparameters.} We evaluate BIRD-Dev and Spider-Dev under a 200-example setting with $K=16$ candidate samples and two backbones, Qwen3-4B and SQL-R1-7B. Bubbles summarize repeated BIRD--Spider accuracy pairs, and curves mark model-specific Pareto frontiers.}
    \label{fig:sensitivity_main}
\end{figure}

We further analyze the sensitivity of \framework{} to its Co-Evolution hyperparameters by sweeping critic variants, memory payload size, evolution depth, consistency weighting, confidence thresholds, time discounting, temperature, and tie-breaking rules on 200-example Spider-Dev and BIRD-Dev subsets. The resulting accuracy pairs form compact clusters rather than isolated optima, suggesting that the gains of \framework{} are not driven by a fragile hyperparameter recipe: many substantially different settings converge to comparable Spider--BIRD trade-offs. More importantly, the non-dominated configurations reveal a structured backbone--dataset interaction. Qwen3-4B tends to push the frontier toward higher Spider-Dev accuracy, whereas SQL-R1-7B attains stronger BIRD-Dev trade-offs, indicating that Co-Evolution mainly reallocates each backbone's existing reasoning capacity rather than imposing a universal optimum. We therefore adopt the default configuration in Section~\ref{setup} and Appendix~\ref{hyperparameter} as a balanced operating point across benchmarks, instead of tuning a separate configuration for each model--dataset pair. Appendix~\ref{app:sensitivity} details more sensitivity analyses for reference.

\subsection{Ablation}

We conduct ablation studies to examine whether Co-Evolution benefits from critic-guided diagnosis and whether an additional Best-of-$K$ selection stage provides reliable improvements over the default greedy utility selection.

\subsubsection{Comparison with Self-Evolution Counterpart}

We evaluate an execution-only Self-Evolution variant that removes LLM-critic diagnosis. Table~\ref{tab:self_evolution_bird} reports the comparison on BIRD-Dev, with full Spider/BIRD results in Appendix~\ref{app:ablation_evolution} and Table~\ref{tab:ablation_self_evolution}. Co-Evolution is generally more reliable, especially on Coder-3B and SQL-R1, while the margin is smaller on Qwen3-4B.

\begin{table}[ht]
    \centering
    \setlength{\tabcolsep}{4pt}
    \caption{BIRD-Dev strict EX~(\%) for Self-Evolution and Co-Evolution. $\Delta$ denotes Co-Evolution minus Self-Evolution.}
    \label{tab:self_evolution_bird}
    \resizebox{\columnwidth}{!}{%
    \begin{tabular}{@{}lccc@{}}
    \toprule
    \textbf{Backbone} & \textbf{Self-Evolution} & \textbf{Co-Evolution} & \textbf{$\Delta$} \\
    \midrule
    Coder-3B & 54.30 & 60.43 & \percentcell{+6.13} \\
    Coder-7B & 60.10 & 63.04 & \percentcell{+2.94} \\
    Qwen3-4B & 66.43 & 66.56 & \percentcell{+0.13} \\
    SQL-R1 & 64.15 & 67.14 & \percentcell{+2.99} \\
    \bottomrule
    \end{tabular}%
    }
\end{table}

\subsubsection{\texorpdfstring{Add-on Best-of-$K$ selector}{Add-on Best-of-K selector}}

Table~\ref{tab:best_of_k_selector} shows that adding a final Best-of-$K$ selector yields only marginal and inconsistent changes over greedy utility selection. This suggests that most of the usable credit-assignment signal has already been captured during Co-Evolution.

\begin{table}[ht]
    \centering
    \setlength{\tabcolsep}{4pt}
    \caption{Strict EX~(\%) of the add-on Best-of-$K$ selector. $\Delta$ is relative to greedy utility selection.}
    \label{tab:best_of_k_selector}
    \begin{tabular}{@{}lccc@{}}
    \toprule
    \textbf{Backbone} & \textbf{Selector} & \textbf{Spider-Test} & \textbf{BIRD-Dev} \\
    \midrule
    \multirow{3}{*}{Coder-3B} & Greedy & 76.99 & 60.43 \\
        & Best-of-$K$ & 77.22 & 60.69 \\
        & $\Delta$ & \percentcell{+0.23} & \percentcell{+0.26} \\
        \midrule
    \multirow{3}{*}{Coder-7B} & Greedy & 81.32 & 63.04 \\
        & Best-of-$K$ & 80.67 & 63.17 \\
        & $\Delta$ & \percentcell{-0.65} & \percentcell{+0.13} \\
        \midrule
    \multirow{3}{*}{Qwen3-4B} & Greedy & 82.95 & 66.56 \\
        & Best-of-$K$ & 83.14 & 67.01 \\
        & $\Delta$ & \percentcell{+0.19} & \percentcell{+0.45} \\
        \midrule
    \multirow{3}{*}{SQL-R1} & Greedy & 84.63 & 67.14 \\
        & Best-of-$K$ & 84.35 & 67.01 \\
        & $\Delta$ & \percentcell{-0.28} & \percentcell{-0.13} \\
    \bottomrule
    \end{tabular}%
\end{table}

\section{Conclusion}

We introduced \framework{}, a Co-Evolution framework for Text2SQL that integrates generation, execution, critique, memory, and utility-based selection into an iterative refinement loop. 
Overall, \framework{} consistently outperforms strong Maj@16 baselines on Spider-Dev, Spider-Test, and BIRD-Dev across multiple open-source backbones. The round-wise results demonstrate clear benefits from iterative evolution: the utility signal identifies high-quality candidates early, and later rounds further correct errors left by the initial generation. The sensitivity analysis shows that the framework remains robust under different hyperparameter settings. 
Ablations further show that critic-guided Co-Evolution is more reliable than execution-only Self-Evolution. 
Overall, these findings suggest that \framework{} offers a practical inference-time scaling approach for Text-to-SQL by combining execution-grounded feedback with contextualized candidate memory, improving generation reliability while better leveraging the native strengths of different backbones.

\section*{Limitations}
\label{main_limit}
Despite the progress made by \framework{} on Co-Evolution for NL2SQL, this study still has the following limitations:

\paragraph{Bounded test-time evolution.}
Although \framework{} performs multi-round refinement, its evolution is still largely problem-local. The candidate memory is maintained for each input instance and is reset after inference, which means that errors, useful critiques, or schema-specific repair patterns discovered on earlier queries are not persistently reused by later queries. This design avoids uncontrolled memory growth and reduces context pressure, but it also limits \framework{} from becoming a truly online-evolving Text2SQL system. Moreover, the memory--retrieval--selection loop is inherently discrete and non-differentiable. As a result, directly applying RLVR-style end-to-end optimization to the full system is difficult: execution, critique, memory update, and utility-based selection must be optimized in stages, which may be suboptimal and increases engineering complexity.

\paragraph{Critic calibration and credit assignment.}
A second bottleneck lies in critic initialization and score calibration. \framework{} relies on the critic to assign useful credit to partially correct or repairable candidates, but cold-start critics, especially smaller LLMs, can produce poorly calibrated scores in early rounds. Miscalibrated feedback can destabilize candidate filtering, prematurely collapse population diversity, or over-rank syntactically plausible but semantically wrong SQL programs. We explored several critic-driven population sampling and re-ranking strategies, but the benefit is not always monotonic: once the group critic and utility function have already extracted the reliable signal from a small candidate pool, additional critic-based selection can mostly re-rank noisy candidates rather than provide new supervision. This suggests that stronger critic pretraining, explicit calibration, or uncertainty-aware scoring may be necessary for more reliable Co-Evolution.

\paragraph{Optimization cost and benchmark saturation.}
Finally, scaling the full refinement harness is computationally expensive. A heavier environment with richer execution traces, external tools, MCP-style interaction, or detailed SQL debugging harnesses may provide stronger supervision, but it also substantially increases token and execution cost; in our preliminary trials, harness-based evaluation can require on the order of billions of tokens per thousand candidate programs. This cost makes large-scale end-to-end optimization difficult. In addition, on stronger recent backbones such as Qwen3-series models, additional fine-tuning on common Text2SQL benchmarks often yields only marginal gains, likely because these models have already absorbed substantial benchmark-adjacent SQL ability. Future work should therefore focus on persistent cross-query memory, calibrated lightweight critics, and lower-cost training signals that can improve difficult out-of-distribution schemas without relying on prohibitively heavy harnesses.

\bibliography{custom}

\newpage
\onecolumn
\appendix
\section{\framework{} settings}

\subsection{Online Co-evolution Algorithm}
\label{app:coevolution_algorithm}

\begin{algorithm}[ht]
\small
\caption{Online co-evolution inference in \framework{}}
\label{alg:coevosql_online}
\begin{algorithmic}[1]
\Require Question $x$, schema $S$, database $D$; generator $M_{\theta}$; critic $M_{\phi}$;
candidate number $K$; maximum rounds $T$; payload size $M$;
time discount $\gamma$; consistency weight $\lambda_{\mathrm{cons}}$;
confidence threshold $\tau$; softmax temperature $\kappa$;
early-stop threshold $\pi_{\mathrm{cons}}$.
\Ensure Final SQL prediction $\hat{q}$.

\State $\mathcal{Q}_{\leq -1} \gets \emptyset$ \Comment{Problem-local episodic memory.}
\State $H_{-1} \gets \emptyset$ \Comment{Cumulative critic summary.}
\State $U_{-1} \gets \emptyset$ \Comment{No scored candidates initially.}

\For{$t = 0, 1, \ldots, T-1$}

    \If{$t = 0$}
        \State $\mathcal{C}_{t} \gets \textsc{Elicit}(M_{\theta}, x, S, D, K)$
        \Comment{Sample $K$ initial SQL candidates with empty payload.}
    \Else
        \For{$i = 1, \ldots, K$}
            \State $\mathcal{P}_{(i,t)} \gets
            \textsc{SoftmaxPayload}(\mathcal{Q}_{\leq t-1}, U_{t-1}, M, \kappa)$
            \Comment{Sample $M$ memory entries as repair payload.}
            \State $q_{(i,t)} \gets
            \textsc{Aggregate}(M_{\theta}, x, S, D, \mathcal{P}_{(i,t)}, H_{t-1})$
        \EndFor
        \State $\mathcal{C}_{t} \gets \{q_{(i,t)}\}_{i=1}^{K}$
    \EndIf
    \ForAll{$q_{(i,t)} \in \mathcal{C}_{t}$}
        \State $e^{d}_{(i,t)} \gets \mathrm{Exec}(q_{(i,t)} \mid D,S)$
        \Comment{Execution status, result, and error log.}
    \EndFor

    \If{$t=T-1$}
        \State $\mathcal{Q}_{\leq t} \gets \textsc{UpdateExec}(\mathcal{Q}_{\leq t-1},\mathcal{C}_t,\{e^d_{(i,t)}\}_{i=1}^{K})$
        \State $\hat{q} \gets \textsc{TerminalSelect}(\mathcal{Q}_{\leq t},U_{t-1})$
        \Comment{Select by execution cluster; no terminal critic call.}
        \State \Return $\hat{q}$
    \EndIf

    \State $E^{p}_{t} \gets
    \textsc{Critic}(M_{\phi}, x, S, D, \mathcal{C}_{t},
    \{e^{d}_{(i,t)}\}_{i=1}^{K}, H_{t-1})$
    \Comment{Parsed critic output for all candidates.}
    \State $H_t \gets \textsc{ParseSummary}(E^{p}_{t})$
    \State $\mathcal{Q}_{\leq t} \gets \mathcal{Q}_{\leq t-1}$

    \ForAll{$q_{(i,t)} \in \mathcal{C}_{t}$}
        \State $e^{p}_{(i,t)} \gets \textsc{ParseEval}(E^{p}_{t}, i)$
        \State $c_{(i,t)} \gets \textsc{ParseConfidence}(e^{p}_{(i,t)})$
        \State $\mathcal{Q}_{\leq t} \gets
        \textsc{Update}(\mathcal{Q}_{\leq t},
        (q_{(i,t)}, e^{d}_{(i,t)}, e^{p}_{(i,t)}, c_{(i,t)}, t))$
        \Comment{Update memory with the latest candidate record.}
    \EndFor

    \ForAll{$q \in \mathcal{Q}_{\leq t}$}
        \State $N_t(q) \gets
        \left|\{q' \in \mathcal{Q}_{\leq t} :
        e^{d}(q') = e^{d}(q),\; \mathrm{Conf}(q') \geq \tau\}\right|$
        \State $\mathrm{Cons}_t(q) \gets \log(1 + N_t(q))$
        \State $U_t(q) \gets
        \gamma^{\,t-t(q)} \cdot \mathrm{Conf}(q)
        + \lambda_{\mathrm{cons}} \cdot \mathrm{Cons}_t(q)$
        \Comment{Time-discounted utility with consistency bonus.}
    \EndFor

    \State $\mathcal{B}_t \gets \mathrm{TopK}(\mathcal{Q}_{\leq t}; U_t, K)$
    \State $r_t^{\star} \gets
    \arg\max_{r}
    \left|\{q \in \mathcal{B}_t : e^{d}(q)=r,\; \mathrm{Conf}(q)\geq \tau\}\right|$

    \State $\rho^{\mathrm{cons}}_t \gets
    \frac{
    \left|\{q \in \mathcal{B}_t : e^{d}(q)=r_t^{\star},\; \mathrm{Conf}(q)\geq \tau\}\right|
    }{|\mathcal{B}_t|}$

    \If{$\rho^{\mathrm{cons}}_t \geq \pi_{\mathrm{cons}}$}
        \State \textbf{break}
        \Comment{Early stop: candidate pool is execution-stable.}
    \EndIf

\EndFor

\State $\hat{q} \gets \textsc{GreedySelect}(\mathrm{TopK}(\mathcal{Q}_{\leq t}; U_t))$
\Comment{Select the highest-utility SQL from the dominant execution cluster.}
\State \Return $\hat{q}$

\end{algorithmic}
\end{algorithm}

\paragraph{Description.}
Algorithm~\ref {alg:coevosql_online} summarizes the online co-evolution procedure used by \framework{}.
Given a natural-language question, database schema, and database instance, the algorithm maintains a
problem-local episodic memory of SQL candidates. The \textsc{Elicit} operator initializes the candidate
population by sampling independent SQL rollouts from the generator. The \textsc{Critic} operator performs
two-stage verification: deterministic execution first records execution status, result, and error messages,
and an LLM critic then produces rubric scores, error diagnoses, and mutation hints. The \textsc{Utility}
operator converts critic scores and execution feedback into calibrated candidate utilities, where invalid
programs are suppressed, empty-result programs are down-weighted, stale candidates are discounted, and
execution-consistent candidates receive a majority-style bonus. Before each non-initial round, the
\textsc{SoftmaxPayload} operator samples memory entries according to a softmax distribution over utilities;
these entries form the few-shot repair payload for the \textsc{Aggregate} operator, which synthesizes the next
candidate population from selected candidates, execution logs, critic diagnoses, and the cumulative summary.
The loop terminates when the top candidates become sufficiently execution-consistent, or when
the maximum round budget is reached. At the terminal budget, \textsc{TerminalSelect} forms execution-result clusters over all accumulated candidates, returns the highest-utility scored member of the dominant cluster when one exists, and otherwise returns the earliest member of a terminal-only cluster.

\begin{table}[ht]
  \centering
  \caption{Co-evolving and memory-related hyperparameters of \framework.}
  \label{tab:self_evolving_hparams}
  \begin{tabular}{ll}
  \toprule
  \textbf{Hyperparameter} & \textbf{Value / Description} \\
  \midrule
   Candidates per round $K$ & 16 \\
   Maximum rounds $T$ & 3 \\
   Payload size $M$ & 4 \\
   Critic mode & Grouped critic w/ \texttt{1\_score} utility prompt \\
   Time discount $\gamma$ & 0.9 \\
   Consistency coefficient $\lambda_{\mathrm{cons}}$ & 0.3 \\
   Confidence threshold $\tau$ & 8.0 \\
   Sampling temperature $\kappa$ & 1.0 \\
   Early-stop consistency ratio $\pi_{\mathrm{cons}}$ & 0.90 \\
  \bottomrule
  \end{tabular}
\end{table}

  \begin{table}[ht]
  \centering
  \caption{Training hyperparameters of \framework\ (SDPO).}
  \label{tab:training_hparams}
  \begin{tabular}{ll}
  \toprule
   \textbf{Hyperparameter} & \textbf{Value / Description} \\
  \midrule
   SDPO stage learning rate & $1\times10^{-5}$ \\
   Training epochs & 5 \\
   Rollouts per question & 8 \\
   Policy loss & SDPO \\
   Distillation granularity & Top-$K$ logit-level distillation \\
   Distillation objective & KL divergence \\
   Top-$K$ distillation & 100 with tail bucket \\
   Teacher regularization & EMA teacher \\
   EMA update rate & 0.05 \\
   Optimizer & AdamW \\
   Weight decay & 0.01 \\
   Warmup ratio & 0.0 \\
   Gradient clipping & 1.0 \\
   PPO mini-batch size & 48 \\
   PPO micro-batch size per GPU & 2 \\
  \bottomrule
  \end{tabular}
  \end{table}

\subsection{Co-evolution hyperparameters}
\label{hyperparameter}
Table~\ref{tab:self_evolving_hparams} summarizes the inference-time hyperparameters used by \framework{}. Unless otherwise specified, all backbone models use the same configuration. We sample $K=16$ SQL candidates per round and run at most $T=3$ co-evolution rounds. The memory payload size is set to $M=4$, so each refinement step conditions on the most useful retrieved candidate summary rather than a large prompt payload. Candidate utility is computed with time discounting and consistency bonus. The early-stop threshold terminates refinement when the top candidate set becomes execution-stable. This configuration is chosen as a balanced setting between search depth, prompt cost, and refinement reliability.

\subsection{SDPO training recipes}
Table~\ref{tab:training_hparams} reports the hyperparameters for the additional SQL-oriented SDPO training stage. This stage is applied only to selected base backbones before running inference-time co-evolution. We use rollouts from the policy model and optimize an SDPO objective with GRPO-style advantage estimation. The distillation signal is applied at the top-$K$ logit level, with a tail bucket to preserve probability mass outside the retained logits. An EMA teacher is used for regularization, stabilizing the privileged execution-aware supervision during training. All other optimization settings, including AdamW, weight decay, gradient clipping, and batch sizes, are kept fixed across the SDPO runs.

\section{Case Study}

\subsection{Co-Evolve Finds the Correct SQL Through Iteration}
\label{case_study_coevolve_iteration}

We present a BIRD case study in which \framework{} recovers the correct SQL only through iterative test-time co-evolution. The generator is \texttt{Qwen2.5-Coder-3B}, the critic is \texttt{Qwen3-4B}, and each round samples $K=16$ candidates with branch count $M=4$ for at most $T=3$ rounds. The critic uses the group four-score setting and evaluates the candidates jointly; execution-equivalence clustering is not used in this example. We report the answer selected greedily by the utility score at each round.

The question asks: \emph{``Please specify all of the schools and their related mailing zip codes that are under Avetik Atoian's administration.''} The gold query is a direct filter over the \texttt{schools} table:

\begin{responsebox}{Gold SQL}
\begin{lstlisting}[language=SQL]
SELECT School, MailZip FROM schools
WHERE AdmFName1 = 'Avetik' AND AdmLName1 = 'Atoian';
\end{lstlisting}
\end{responsebox}

This case is challenging for single-round selection because the initial candidate pool contains no correct SQL. As shown in Table~\ref{tab:case_study_multiround}, all 16 Round~0 candidates fail execution. The SQL trace below further shows that the greedy top-utility candidate hallucinates a table named \texttt{adm\_acct}. Since the correct program is absent from the first candidate set, neither majority voting nor reranking the Round~0 pool can recover the answer.

\begin{table}[ht]
    \centering
    \setlength{\tabcolsep}{4pt}
    \caption{BIRD case study for multi-round co-evolution. Buckets are reported as invalid/low/high counts: $-1$ denotes invalid or failed candidates, $1$--$4$ denote low-confidence candidates, and $10$ denotes high-confidence candidates. For this example, utility and confidence bucket counts are identical. The cumulative column reports unique SQL / valid candidates / correct candidates after each round.}
    \label{tab:case_study_multiround}
    \begin{tabular}{@{}ccccccp{2.2cm}@{}}
    \toprule
    \textbf{Round} & \textbf{Utility} & \textbf{Conf.} & \textbf{Correct} & \textbf{Valid} & \textbf{Buckets} & \textbf{Cumulative pool} \\
    \midrule
    0 & $-1.0000$ & $-1.0$ & 0 & 0 & 16 / 0 / 0 & 16 / 0 / 0 \\
    1 & $5.0497$ & $10.0$ & 11 & 12 & 4 / 1 / 11 & 26 / 12 / 11 \\
    2 & $5.0652$ & $10.0$ & 14 & 14 & 2 / 0 / 14 & 31 / 26 / 25 \\
    \bottomrule
    \end{tabular}
\end{table}

\begin{responsebox}{Case Study Example Extracted from Bird Dataset (Problem Index 54)}
\begin{lstlisting}[language=SQL]
-- Round 0: invalid top-utility candidate
SELECT s.School, s.MailZip
FROM schools AS s
JOIN adm_acct AS ac ON s.CDSCode = ac.cds
WHERE ac.adm_fname1 = 'Avetik' AND ac.adm_lname1 = 'Atoian';
-- execution: no such table: adm_acct
-- utility = -1.0000, confidence = -1.0, correct = 0 / 16

-- Round 1: correct SQL first appears after critic feedback
SELECT DISTINCT s.School, s.MailZip
FROM schools s
WHERE s.AdmFName1 = 'Avetik' AND s.AdmLName1 = 'Atoian';
-- result: [['Fairview Middle', '93926-0238']]
-- utility = 5.0497, confidence = 10.0, correct = 11 / 16

-- Round 2: cleaner gold-equivalent SQL
SELECT School, MailZip FROM schools
WHERE AdmFName1 = 'Avetik' AND AdmLName1 = 'Atoian';
-- result: [['Fairview Middle', '93926-0238']]
-- utility = 5.0652, confidence = 10.0, correct = 14 / 16
\end{lstlisting}
\end{responsebox}

The SQL trace above illustrates how the critic-guided loop changes the candidate distribution rather than merely reordering a fixed set. In Round~0, the selected SQL fails because it joins against the non-existent \texttt{adm\_acct} table. The critic identifies this schema error and points to the administrator fields already present in \texttt{schools}, namely \texttt{AdmFName1} and \texttt{AdmLName1}. With this feedback, Round~1 generates a correct query that filters \texttt{schools} by the administrator's first and last name and returns the required columns, \texttt{School} and \texttt{MailZip}. Round~2 further simplifies the answer by removing unnecessary aliasing and \texttt{DISTINCT}, yielding the gold-equivalent SQL.

The key observation is that \framework{} succeeds because it creates new, better candidates after receiving structured criticism. The correct SQL is unavailable in Round~0, appears in Round~1, and becomes dominant by Round~2: the number of correct candidates increases from $0/16$ to $11/16$ and then to $14/16$, while the accumulated pool contains 25 correct candidates out of 48 total candidates by the end of the trace. This demonstrates that co-evolution can turn an initially unusable candidate set into a high-confidence solution through iterative, execution-grounded repair.

\section{Additional Experiments}
\label{additional_results_and_analysis}

\subsection{Analysis of Single-Turn Models}

Table~\ref{tab:single_turn_models} summarizes the single-turn performance of representative backbones before multi-round evolution. With $K=16$ samples, Avg@16 measures the mean execution accuracy over all rollouts, while All@16 requires all rollouts to be correct and thus reflects stability. Pass@16 is an oracle-style metric that requires at least one correct candidate among the 16 samples. Maj@16 measures whether majority aggregation can recover the correct SQL from the sampled set.

The results show that Qwen3-4B is highly stable and even surpasses the GRPO-trained Qwen2.5-Coder-7B variant (SQL-R1) on several single-turn metrics, suggesting that the capability of the base model plays a critical role in subsequent test-time scaling.

\begin{table}[ht]
    \centering
    \caption{Single-turn execution accuracy (\%) of representative backbone models under no thinking mode. Results are reported with 16 sampled candidates.}
    \label{tab:single_turn_models}
    \setlength{\tabcolsep}{6pt}
    \begin{tabular}{llccc}
    \toprule
    \textbf{Metric} & \textbf{Backbone} & \textbf{Spider-Dev} & \textbf{Spider-Test} & \textbf{BIRD-Dev} \\
    \midrule
  \multirow{4}{*}{\textbf{Avg@16}}
      & Coder-3B & 56.62 & 58.25 & 28.15 \\
      & Coder-7B & 73.34 & 75.01 & 46.99 \\
      & Qwen3-4B & 82.44 & 80.80 & 62.12 \\
      & SQL-R1 & 81.33 & 82.57 & 61.51 \\
  \midrule
  \multirow{4}{*}{\textbf{All@16}}
      & Coder-3B & 10.74 & 13.65 & 1.43 \\
      & Coder-7B & 42.55 & 42.85 & 14.67 \\
      & Qwen3-4B & 70.31 & 69.68 & 47.07 \\
      & SQL-R1 & 67.41 & 67.96 & 38.98 \\
  \midrule
  \multirow{4}{*}{\textbf{Maj@16}}
      & Coder-3B & 76.40 & 76.01 & 51.24 \\
      & Coder-7B & 80.08 & 81.23 & 61.60 \\
      & Qwen3-4B & 84.53 & 82.35 & 65.19 \\
      & SQL-R1 & 83.08 & 83.74 & 65.65 \\
  \midrule
  \multirow{4}{*}{\textbf{Pass@16}}
      & Coder-3B & 88.59 & 89.89 & 65.91 \\
      & Coder-7B & 90.91 & 91.06 & 75.23 \\
      & Qwen3-4B & 89.94 & 87.98 & 73.73 \\
      & SQL-R1 & 90.72 & 91.66 & 77.71 \\
    \bottomrule
    \end{tabular}
\end{table}

For each pair, we also run one independent $K=48$ pool to match the maximum equivalent computing resources of \framework{}. Table \ref{tab:matched_parallel_sampling} shows that Maj@32 to Maj@48 adds only 0.17 EX on average; \framework{} exceeds Maj@48 in 7/12 settings, including all four BIRD backbones, while Majority Voting is competitive in several Spider settings.

\begin{table*}[ht]
    \centering
    \setlength{\tabcolsep}{4pt}
    \caption{Strict EX~(\%) under different Majority-voting sampling budgets.
    Deltas denote \framework{} (Base) minus the corresponding Maj@$K$.}
    \label{tab:matched_parallel_sampling}
    \begin{tabular}{@{}llrrrrrr@{}}
    \toprule
    \textbf{Dataset} & \textbf{Backbone}
    & \textbf{Maj@16} & \textbf{Maj@32} & \textbf{Maj@48}
    & \textbf{$\Delta$ vs. 16} & \textbf{$\Delta$ vs. 32} & \textbf{$\Delta$ vs. 48} \\
    \midrule
    \multirow{4}{*}{BIRD-Dev}
      & Coder-3B & 50.98 & 53.00 & 53.78
      & \percentcell{+9.45} & \percentcell{+7.43} & \percentcell{+6.65} \\
      & Coder-7B & 62.13 & 61.99 & 62.06
      & \percentcell{+0.91} & \percentcell{+1.05} & \percentcell{+0.98} \\
      & Qwen3-4B & 65.12 & 64.80 & 65.12
      & \percentcell{+1.44} & \percentcell{+1.76} & \percentcell{+1.44} \\
      & SQL-R1 & 65.19 & 65.65 & 66.04
      & \percentcell{+1.95} & \percentcell{+1.49} & \percentcell{+1.10} \\
    \midrule
    \multirow{4}{*}{Spider-Dev}
      & Coder-3B & 76.69 & 77.66 & 77.47
      & \percentcell{+1.16} & \percentcell{+0.19} & \percentcell{+0.38} \\
      & Coder-7B & 80.56 & 81.33 & 81.72
      & \percentcell{+0.97} & \percentcell{+0.20} & \percentcell{-0.19} \\
      & Qwen3-4B & 84.62 & 85.01 & 84.91
      & \percentcell{+0.20} & \percentcell{-0.19} & \percentcell{-0.09} \\
      & SQL-R1 & 83.37 & 84.14 & 83.85
      & \percentcell{+0.09} & \percentcell{-0.68} & \percentcell{-0.39} \\
    \midrule
    \multirow{4}{*}{Spider-Test}
      & Coder-3B & 76.43 & 76.99 & 77.41
      & \percentcell{+0.56} & \percentcell{+0.00} & \percentcell{-0.42} \\
      & Coder-7B & 82.63 & 82.77 & 82.72
      & \percentcell{-1.31} & \percentcell{-1.45} & \percentcell{-1.40} \\
      & Qwen3-4B & 82.44 & 82.58 & 82.77
      & \percentcell{+0.51} & \percentcell{+0.37} & \percentcell{+0.18} \\
      & SQL-R1 & 83.56 & 83.70 & 83.79
      & \percentcell{+1.07} & \percentcell{+0.93} & \percentcell{+0.84} \\
    \bottomrule
    \end{tabular}
\end{table*}

\subsection{Complete SQL Benchmark Results}
\label{app:full_sql_bench}

Table ~\ref{tab:full_sql_bench_relax} reports the complete SQL benchmark results under relaxed execution accuracy. Here, \textbf{Strict} directly compares the execution result of the predicted SQL against the gold result. \textbf{Relaxed} uses the same execution criterion but ignores column order while preserving row order. This relaxation is useful because permuting selected columns usually does not change the semantic answer, whereas row order can be part of the intended result when the query contains ordering modifiers such as \texttt{ORDER BY} or \texttt{DESC}.

\begin{table}[ht]
    \centering
    \caption{Relaxed execution accuracy~(\%) on Spider and BIRD benchmarks.}
    \label{tab:full_sql_bench_relax}
    \begin{tabular}{clcccccc}
    \toprule
        \textbf{Setting} & \textbf{Backbone} & \multicolumn{2}{c}{\textbf{Spider-Dev}} & \multicolumn{2}{c}{\textbf{Spider-Test}} & \multicolumn{2}{c}{\textbf{BIRD-Dev}} \\
        \cmidrule(lr){3-4} \cmidrule(lr){5-6} \cmidrule(lr){7-8}
        & & \textbf{Acc.} & \textbf{$\Delta$} & \textbf{Acc.} & \textbf{$\Delta$} & \textbf{Acc.} & \textbf{$\Delta$} \\
    \midrule
    \multirow{4}{*}{\textbf{Maj@16}}
          & Coder-3B & 81.14 & -- & 80.39 & -- & 51.30 & -- \\
          & Coder-7B & 84.53 & -- & 85.93 & -- & 61.86 & -- \\
          & Qwen3-4B & 87.33 & -- & 87.19 & -- & 65.38 & -- \\
          & SQL-R1 & 87.04 & -- & 88.50 & -- & 66.04 & -- \\
    \midrule
    \multirow{4}{*}{\shortstack{\textbf{\framework}\\(Base)}}
        & Coder-3B & 82.88 & \percentcell{+1.74} & 81.51 & \percentcell{+1.12} & 61.08 & \percentcell{+9.78} \\
        & Coder-7B & 85.78 & \percentcell{+1.25} & 85.98 & \percentcell{+0.05} & 63.30 & \percentcell{+1.44} \\
        & Qwen3-4B & 87.62 & \percentcell{+0.29} & 87.89 & \percentcell{+0.70} & 66.82 & \percentcell{+1.44} \\
        & SQL-R1 & 87.04 & \percentcell{+0.00} & 89.10 & \percentcell{+0.60} & 67.47 & \percentcell{+1.43} \\
    \midrule
    \multirow{2}{*}{\shortstack{\textbf{\framework}\\(SDPO)}}
          & Coder-3B & 80.95 & \percentcell{-0.19} & 83.23 & \percentcell{+2.84} & 61.28 & \percentcell{+9.98} \\
          & Qwen3-4B & 86.17 & \percentcell{-1.16} & 87.80 & \percentcell{+0.61} & 67.28 & \percentcell{+1.90}  \\
    \bottomrule
    \end{tabular}
\end{table}

\subsection{Sensitivity on \framework{}}
\label{app:sensitivity}

\begin{figure}[ht]
    \centering
    \includegraphics[width=\linewidth]{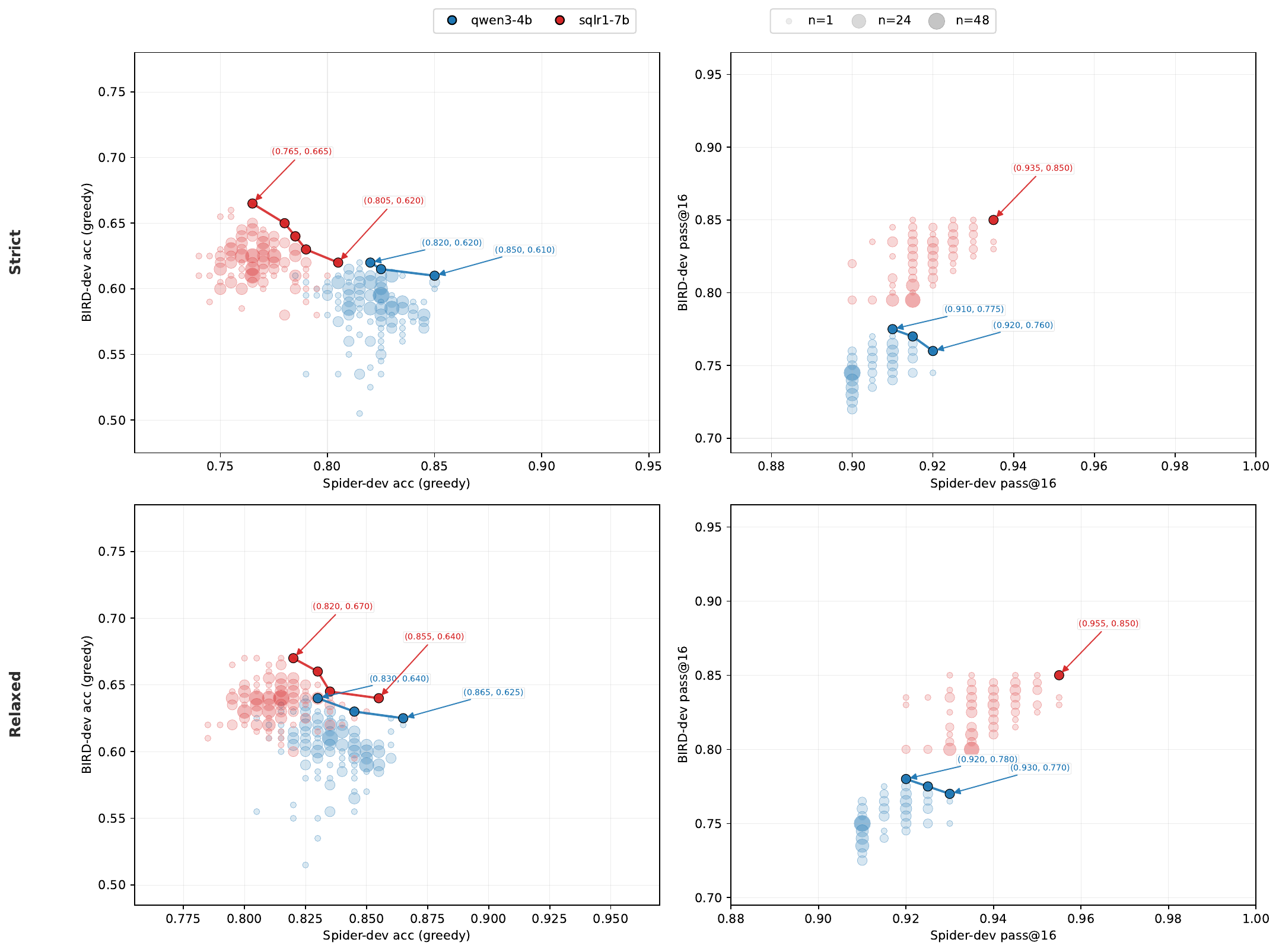}
    \caption{\textbf{Sensitivity Analysis of Co-Evolution Hyperparameters.} We evaluate BIRD-Dev and Spider-Dev under a 200-example setting with $K=16$ candidate samples and two backbones, Qwen3-4B and SQL-R1-7B. Bubbles summarize repeated BIRD--Spider accuracy pairs, and curves mark model-specific Pareto frontiers across strict/relaxed accuracy and greedy/Pass@16 protocols.}
    \label{fig:sensitivity}
\end{figure}

Figure~\ref{fig:sensitivity} visualizes a broad ablation grid over group and no-group critic variants, 1-score and 4-score utility prompts, payload sizes $M \in \{1,4\}$, iteration depths $T \in \{1,2,3\}$, consistency weights $\{0.0,0.1,0.3\}$, confidence thresholds $\tau \in \{8,10\}$, time discounts $\gamma \in \{0.9,1.0\}$, temperature variants $\{0.5,0.7\}$,  tie-breaking rules, consistency-bonus caps, and checklist-style critic prompts. In total, the ablation catalog contains 110 variant specifications with 372 paired BIRD-Spider configuration points for each evaluation protocol. Rows compare strict execution accuracy and relaxed execution accuracy, while columns report greedy accuracy and Pass@16. Each bubble corresponds to a unique Spider-Dev accuracy and BIRD-Dev accuracy pair; bubble size and opacity indicate how many hyperparameter configurations collapse to the same pair, making the background a discrete density summary rather than a set of independent repeated marks. The foreground curves show the Pareto structure of the sensitivity sweep.

The sweep suggests some practical observations. First, not every nominal hyperparameter acts as an effective degree of freedom: many different ablation settings map to the same bubble, indicating that several design choices are robust within the evaluated range. Second, the preferred setting depends on both the backbone model and the target dataset. Qwen3-4B provides stronger Spider-Dev trade-off regions, while SQL-R1-7B provides stronger BIRD-Dev frontiers, suggesting that co-evolution policies can benefit from model- or dataset-specific calibration.

\subsection{Ablation on Evolution Method}
\label{app:ablation_evolution}

\begin{figure}[ht]
    \centering
    \includegraphics[width=\linewidth]{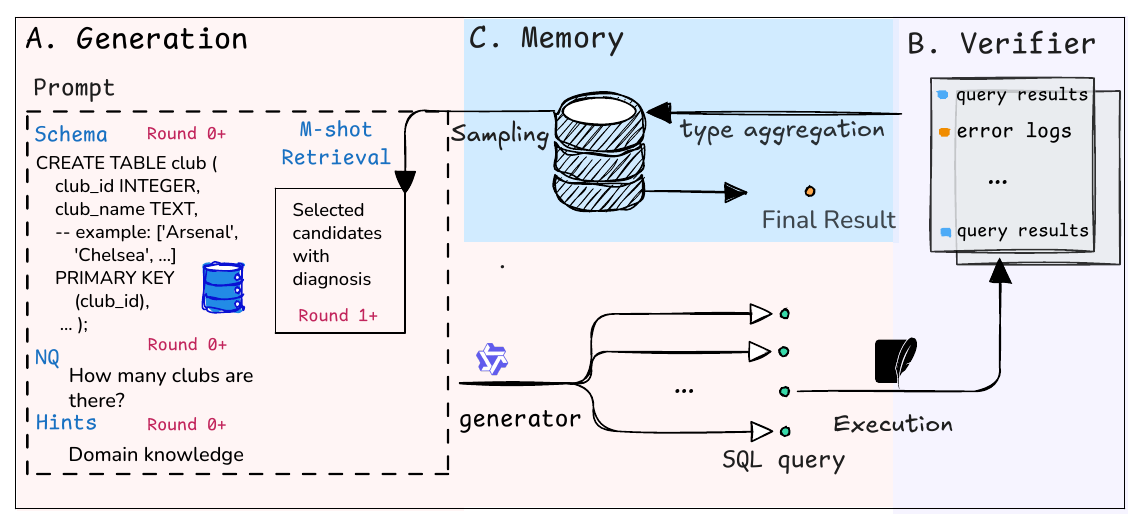}
    \caption{\textbf{Self-Evolution Counterpart.} Given the question and schema, the Self-Evolution method initializes multiple role-based groups to generate SQL candidates in parallel, followed only by execution-based verification. Candidates are stored in a candidate pool, retrieved based on execution type, and the final answer is selected greedily or an additional Best-of-$K$ selector.}
    \label{fig:rollout}
\end{figure}

To isolate the contribution of the LLM-based critic, we consider a \emph{Self-Evolution} variant that removes critic scoring and textual diagnosis, and lets the generator carry out the refinement process using only its own generated candidates and deterministic execution feedback. As illustrated in Figure~\ref{fig:rollout}, this variant preserves the multi-round candidate memory and execution-based verification, but replaces critic-guided mutation with generator-driven self-refinement. The final answer is selected greedily by utility from the Self-Evolved candidate pool. Table~\ref{tab:ablation_self_evolution} reports the corresponding Self-Evolution results on Spider and BIRD.

\begin{table*}[ht]
    \caption{Greedy strict EX~(\%) of the Self-Evolution counterpart on Spider and BIRD benchmarks.}
    \centering
    \begin{tabular}{lccc}
    \toprule
      \textbf{Backbone} & \textbf{Spider-Dev} & \textbf{Spider-Test} & \textbf{BIRD-Dev}  \\
    \midrule
        Coder-3B & 73.21 & 74.20 & 54.30 \\
        Coder-7B & 76.50 & 79.13 & 60.10 \\
        Qwen3-4B & 80.17 & 83.79 & 66.43 \\
        SQL-R1 & 80.85 & 83.42 & 64.15 \\
    \bottomrule
    \end{tabular}
    \label{tab:ablation_self_evolution}
\end{table*}
Overall, Self-Evolution recovers part of the benefit from multi-candidate inference, but it remains less stable than critic-guided co-evolution because it lacks explicit LLM-critic diagnosis.

\subsection{Analysis and Comparison of Efficiency}
\label{analysis_efficiency}

We further compare the inference efficiency of critic-guided co-evolution and the Self-Evolution counterpart on the full BIRD development set with 1,534 problems and four backbones: Coder-3B, Coder-7B, Qwen3-4B, and SQL-R1. We exclude the optional Best-of-$K$ add-on here. For clean token accounting, we run the 1-score critic version; the 4-score counterpart has nearly identical early-stop generation volume with a slightly higher parsing-error rate because the JSON schema is relatively more complex.

Table~\ref{tab:bird_efficiency_round_breakdown} gives the full method-level comparison and decomposes token usage by round. Self-Evolution spends 1,071.37M maj@16-equivalent tokens, while co-evolution with critic-guided early stopping spends 753.92M tokens, corresponding to a 29.6\% reduction. Under the logged accounting used by the actual batched run, co-evolution reduces total tokens from 799.38M to 481.94M, a 39.7\% reduction. The critic is therefore not merely additional overhead: its 38.07M-token cost enables early stopping that reduces aggregate-generation cost from 746.23M to 390.77M, saving 355.46M aggregate tokens.

The main difference is in the aggregate rounds: Self-Evolution continues to produce 98,176 aggregate outputs in both round~1 and round~2, whereas co-evolution produces only 51,824 and 44,400 aggregate outputs after critic-guided filtering. Thus co-evolution reduces aggregate generation by
\begin{equation}
    1 - \frac{51{,}824 + 44{,}400}{98{,}176 + 98{,}176} = 51.0\%.
\end{equation}
The critic is shown with the preceding generation round because it scores the current candidate set and supplies the signal used by the next aggregation step. There are two critic rounds: after round~0 candidates and after round~1 candidates.

\begin{table}[ht]
    \centering
    \caption{Round-level token decomposition on BIRD full dev. Values are in millions. Parentheses denote maj@16-equivalent token counts, where the initial \texttt{SamplingParams(n=16)} elicitation prefill is counted once per generated completion rather than once per request. Dashes indicate that the method has no critic call or no subsequent scoring round.}
    \label{tab:bird_efficiency_round_breakdown}
    \resizebox{\textwidth}{!}{%
    \begin{tabular}{lllrrrrr}
    \toprule
    \textbf{Method} & \textbf{Round} & \textbf{Gen. phase} & \textbf{Gen. infill} & \textbf{Gen. decode} & \textbf{Critic infill} & \textbf{Critic decode} & \textbf{Total} \\
    \midrule
    Co-evolve & round 0 & elicit & 18.13 (290.11) & 34.96 (34.96) & 17.22 (17.22) & 3.34 (3.34) & 73.65 (345.63) \\
    Co-evolve & round 1 & aggregate & 194.62 (194.62) & 16.15 (16.15) & 14.72 (14.72) & 2.80 (2.80) & 228.29 (228.29) \\
    Co-evolve & round 2 & aggregate & 166.92 (166.92) & 13.08 (13.08) & -- & -- & 180.01 (180.01) \\
    \textbf{Co-evolve total} & -- & -- & \textbf{379.67 (651.65)} & \textbf{64.19 (64.19)} & \textbf{31.94 (31.94)} & \textbf{6.14 (6.14)} & \textbf{481.94 (753.92)} \\
    \midrule
    Self-Evolve & round 0 & elicit & 18.13 (290.11) & 35.02 (35.02) & -- & -- & 53.15 (325.14) \\
    Self-Evolve & round 1 & aggregate & 333.45 (333.45) & 39.12 (39.12) & -- & -- & 372.58 (372.58) \\
    Self-Evolve & round 2 & aggregate & 334.84 (334.84) & 38.81 (38.81) & -- & -- & 373.65 (373.65) \\
    \textbf{Self-Evolve total} & -- & -- & \textbf{686.42 (958.41)} & \textbf{112.96 (112.96)} & -- & -- & \textbf{799.38 (1071.37)} \\
    \bottomrule
    \end{tabular}}
\end{table}

Table~\ref{tab:bird_efficiency_early_stop} reports where this reduction comes from. Since each active problem produces 16 aggregate outputs per round, the active-problem counts directly determine later-round generation volume. Self-Evolution has no critic-guided early stop in this run and therefore generates 98,176 aggregate outputs in both round~1 and round~2.

\begin{table}[ht]
    \centering
    \setlength{\tabcolsep}{5pt}
    \caption{Early-stopping efficiency on BIRD full dev. 
    We report the number of active problems in each round, with the percentage 
    stopped relative to Round~0 in parentheses. Each active problem generates 
    16 candidate outputs per round.}
    \label{tab:bird_efficiency_early_stop}
    \begin{tabular}{lrrr}
    \toprule
    \textbf{Model} & \textbf{Round 0} & \textbf{Round 1} & \textbf{Round 2} \\
    \midrule
    Coder-3B  & 1,534 & 1,386 (9.6\%)  & 1,371 (10.6\%) \\
    Coder-7B  & 1,534 &   959 (37.5\%) &   864 (43.7\%) \\
    Qwen3-4B  & 1,534 &   430 (72.0\%) &   253 (83.5\%) \\
    SQL-R1    & 1,534 &   464 (69.8\%) &   287 (81.3\%) \\
    \midrule
    \textbf{Total}
              & \textbf{6,136}
              & \textbf{3,239 (47.2\%)}
              & \textbf{2,775 (54.8\%)} \\
    \bottomrule
    \end{tabular}
\end{table}

The active set shrinks substantially after the first evaluation. Across the four models, co-evolution stops 47.2\% of instances before round~1 and 54.8\% before round~2, directly avoiding subsequent generation for these instances. This effect is particularly pronounced for Qwen3-4B and SQL-R1, where more than 80\% of instances no longer require round~2 generation. These results show that the critic enables adaptive compute allocation: a modest scoring overhead substantially reduces the generation cost of later rounds.

\section{Prompt Templates}
This appendix organizes the prompt templates into three functional modules: (i) the \emph{generator}, which proposes and refines SQL queries; (ii) the \emph{critic}, which evaluates candidate correctness or eliminates wrong execution-equivalent clusters; and (iii) an optional \emph{Best-of-$K$ add-on}, which performs a final listwise selection among top-ranked candidates. Across all modules, the prompts follow the same core principle: the Schema, Question, and Hint are authoritative, while candidate SQL and execution results are treated as useful but fallible evidence. This design keeps the system conservative about schema usage, output columns, filters, aggregation, ordering, limits, and row cardinality.

\subsection{Generator}
The generator module contains the prompts used to create and revise SQL candidates. Its design goal is to make every generation step explicitly grounded in the Schema, Question, and Hint, rather than in memorized SQL patterns or previous candidates. The initial prompt elicits one executable query from scratch, while the aggregation and Self-Evolution prompts use previous candidates only as evidence for possible tables, joins, filters, and aggregations. In all generator variants, the model is instructed to reason about requested columns, row granularity, joins, aggregation, ordering, limits, and numeric scale before emitting exactly one final SQL query.

\subsubsection{\framework{} used in rollout phase}

\begin{promptbox}{Generator Prompt (Round 0 / Elicit)}

\begin{lstlisting}
Schema:
{schema}

Question:
{question}

Hint:
{hint}

You are solving a text-to-SQL task.

Use only the Schema, Question, and Hint.
Treat Question and Hint as separate fields. The Hint is unique to this question and should be used only as question-specific evidence.

Think step by step before writing the SQL. It is useful to consider alternative valid SQL formulations, but the final answer should contain one executable query.

In your reasoning, verify:
- The target entity or rows.
- The exact requested output columns and their order.
- Required filters from the Question or Hint.
- Required joins, preferably using stable ID/key columns.
- Required aggregation, grouping, ordering, HAVING, or LIMIT.
- The requested cardinality: one row, multiple rows, grouped rows, or one aggregate value.
- Whether numeric fields are raw values, fractions, percentages, counts, dates, or IDs.

Rules:
- Do not invent tables, columns, filters, joins, or calculations.
- Do not copy a SQL pattern unless it is supported by this Question and Hint.
- Return only the requested columns, in the requested order.
- Avoid SELECT *.
- Avoid unnecessary DISTINCT.
- Use WHERE for base-row filters and HAVING for aggregate filters.
- Be careful with row cardinality for superlatives and with the scale/formula of rates or percentages.

Output format:
Reasoning:
<brief step-by-step reasoning>

Final SQL:
```sql
SELECT column FROM table WHERE condition;
```
\end{lstlisting}

\end{promptbox}

\begin{promptbox}{Generator Prompt (Round 1+ / Aggregate)}

\begin{lstlisting}
Schema:
{schema}

Question:
{question}

Hint:
{hint}

{summary_section}

## Candidate Cards
Each card may contain both useful evidence and serious mistakes.
The execution summary shows behavior, not semantic correctness.
The critic issue summary is useful but may be incomplete or wrong.
Do not trust a candidate merely because it executed successfully.
Do not merge SQL fragments unless each fragment is supported by the Schema, Question, and Hint.

{candidates_section}

## Task
Produce exactly ONE corrected SQL query.

Think step by step before writing the final SQL.

Reasoning protocol:
1. Re-read the original Question and Hint first.
2. Identify the requested output columns and row granularity.
3. Identify the required tables, joins, filters, aggregation, ordering, and limits.
4. Compare candidates as possible evidence, not as truth.
5. Reject candidate logic that uses unsupported filters, wrong joins, wrong aggregation, wrong ordering, wrong limit, or wrong output columns.
6. Prefer the simplest SQL that directly answers the Question.
7. Prefer a small correction to a well-supported candidate, but rewrite from scratch if all candidates share the same semantic error.

SQL construction constraints:
- Treat the Question and Hint as separate fields.
- Use the Hint only as evidence for this specific question.
- Use schema-defined keys for joins when available.
- Return only the requested columns, in the requested order.
- Match the requested row cardinality and granularity.
- Use ordering, aggregation, grouping, or limits only when required by the question.
- Preserve the numeric scale implied by the schema and question.
- If the question explicitly defines a formula, compute that formula directly.
- Do not add DISTINCT, GROUP BY, HAVING, filters, joins, or extra output columns unless required.

Output format:
Reasoning:
<brief step-by-step reasoning>

Final SQL:
```sql
SELECT column FROM table WHERE condition;
```
\end{lstlisting}

\end{promptbox}

\subsubsection{Self-Evolution variant used in rollout phase}

\begin{promptbox}{Generator Prompt (Round 1+ / Aggregate)}

\begin{lstlisting}
Schema:
{schema}

Question:
{question}

Hint:
{hint}

## Candidate Cards
{card_description}

{candidate_blocks}

## Task
Produce exactly ONE improved SQL query.

Think step by step before writing the final SQL.

Reasoning protocol:
1. Reconstruct the intended answer from the original Question and Hint.
2. Identify the requested output columns and their order.
3. Identify the requested row granularity and result count.
4. Identify which tables, joins, filters, aggregation, grouping, ordering, and limits are actually required.
5. Use candidates only as possible evidence for tables, joins, filters, and aggregations.
6. Do not copy candidate SQL blindly.
7. Keep only candidate components that are directly supported by the Schema, Question, and Hint.
8. If candidates disagree, choose the logic best supported by the original task, not the majority.
9. Prefer a simple, direct query over a complex query with extra assumptions.

SQL checklist:
- Correct output columns and order.
- Correct row granularity.
- Correct filters and value matching.
- Correct joins using schema keys when available.
- Correct aggregation and grouping.
- Correct ordering direction and result count when ordering or limiting is required.
- Correct treatment of rates, percentages, dates, and units.
- No unsupported DISTINCT, GROUP BY, HAVING, filters, calculations, or extra joins.

Output format:
Reasoning:
Briefly explain the intended answer, the useful candidate components, the rejected mistakes, and the final SQL structure.

Final SQL:
```sql
SELECT column FROM table WHERE condition;
```
\end{lstlisting}

\end{promptbox}

\subsection{Critic}

The critic module provides the diagnostic signal for online evolution. Unlike the generator, the critic does not write new SQL; it judges whether existing candidates satisfy the original task under SQLite execution semantics. Its design principles are strictness, repairability, and compact memory: execution success is not treated as proof of correctness, semantic mistakes trigger hard score caps, issue summaries must identify concrete repairs, and cumulative summaries retain only stable task-relevant diagnostics. We use the four-score critic as the default and most frequently used backbone, retain a one-score variant as a lightweight ablation, and include a no-group four-score variant for sensitivity studies that score individual candidates without execution-equivalent clustering.

\paragraph{Scoring mode.}
This mode applies when the critic must assign quality scores. The four-score configuration, with alignment/schema/logic/completeness dimensions, is the default backbone and the most frequently used setting in our experiments. A one-score configuration is retained as a lightweight variant that collapses these dimensions into a single scalar quality score. We also include a no-group scoring variant as an ablation: it reuses the same scoring rules but evaluates individual candidates rather than execution-equivalent cluster representatives. Fields such as \texttt{cumulative\_summary}, \texttt{issues}, and \texttt{execution\_assessment} are returned by default for carrying compact cross-round diagnostics.

\begin{promptbox}{Critic Prompt (Scoring Mode)}

\begin{lstlisting}
Evaluate a batch of SQL candidates as a strict SQLite Text-to-SQL judge.

Schema:
{schema}

Question:
{question}

Hint:
{hint}

## Prior cumulative summary
{prior_cumulative_summary}

Use the prior summary only as compact context. Current SQL, question, hint, schema, and execution evidence are authoritative.

## Candidate SQL and Precomputed Execution Evidence
{mode_note}
{candidates_block}

## Task
For each candidate or cluster representative:
1. classify execution,
2. score four dimensions,
3. write one concrete issue summary.

No reference answer is available.
The SQL has already been executed; do not invent execution results.
Treat candidates independently unless their SQL and execution behavior are equivalent.
When the candidate is wrong, `issues` must include both the problem and a repair hint.

## Dimensions
- alignment: whether the SQL answers the question's metric, entity, filters, and requested outputs.
- schema: whether tables, columns, aliases, literals, and SQLite syntax are valid and appropriate.
- logic: whether joins, filters, grouping, aggregation, arithmetic, ordering, limits, and NULL handling are correct.
- completeness: whether the output columns, order, row granularity, and result shape exactly satisfy the question.

## Scoring
Use integer scores from -1 to 10.
- 10: correct.
- 8-9: likely correct with only harmless style or alias issues.
- 5-7: partially correct but with a meaningful uncertainty or issue.
- 1-4: materially wrong.
- 0: invalid or unusable.
- -1: timeout or unavailable execution.

Caps:
- error: all fields set to {execution_error_score}.
- timeout: all fields set to {execution_timeout_score}.
- valid_empty: all fields set to {execution_valid_empty_score}, unless the question explicitly expects no rows.
- If there is a material semantic issue, at least one of alignment, logic, or completeness must be <= 4.
- Wrong requested output, wrong metric, wrong filter, wrong join, wrong aggregation, wrong grouping level, wrong ordering direction, or wrong result count should cap affected dimensions at <= 4.
- Extra selected columns should cap completeness at <= 4 unless they are explicitly requested.
- If the candidate contradicts the Hint on a formula, filter, entity, metric, or granularity, cap affected dimensions at <= 4.

Judging notes:
- Execution success does not imply semantic correctness.
- Empty results are a strong negative signal unless expected.
- Use result previews as evidence for output shape, duplicated rows, NULLs, scalar/list form, and implausible aggregates.
- Preserve the numeric scale implied by the schema, question, and hint.
- If the question explicitly defines a formula, judge whether the candidate computes or faithfully represents that formula.
- Match the requested row granularity and result cardinality.
- Use ordering, aggregation, grouping, or limits only when required by the question.
- Do not penalize harmless aliasing or formatting.

## Cumulative summary
Return `cumulative_summary`.
Keep only stable, task-relevant information:
- execution pattern,
- recurring issue pattern,
- repair guidance.
Do not include SQL code.
Do not copy long prior summaries.

## Output
Return exactly one valid JSON object and nothing else:
{{
  "cumulative_summary": "<compact updated summary>",
  "evaluations": [
    {{
      "candidate_index": 1,
      "execution_assessment": "valid_nonempty | valid_empty | error | timeout",
      "scores": {{"alignment": <-1..10>, "schema": <-1..10>, "logic": <-1..10>, "completeness": <-1..10>}},
      "issues": "<problem plus concrete fix, or 'none' if correct>"
    }}
  ]
}}

\end{lstlisting}

\end{promptbox}

\begin{promptbox}{Critic Prompt (One-score Scoring Variant)}

\begin{lstlisting}
This one-score variant is identical to the four-score backbone except for the fields shown below.
Unchanged instructions are abbreviated as `...`.
Use `<<<` for the four-score backbone text and `>>>` for the one-score replacement.

...

## Task
For each candidate or cluster representative:
1. classify execution,
<<<
2. score four dimensions,
>>>
2. assign one overall quality score,
3. write one concrete issue summary.

...

<<<
Treat candidates independently unless their SQL and execution behavior are equivalent.

## Dimensions
- alignment: whether the SQL answers the question's metric, entity, filters, and requested outputs.
- schema: whether tables, columns, aliases, literals, and SQLite syntax are valid and appropriate.
- logic: whether joins, filters, grouping, aggregation, arithmetic, ordering, limits, and NULL handling are correct.
- completeness: whether the output columns, order, row granularity, and result shape exactly satisfy the question.

## Scoring
Use integer scores from -1 to 10.
- 10: correct.
- 8-9: likely correct with only harmless style or alias issues.
- 5-7: partially correct but with a meaningful uncertainty or issue.
- 1-4: materially wrong.
- 0: invalid or unusable.
- -1: timeout or unavailable execution.

Caps:
- error: all fields set to {execution_error_score}.
- timeout: all fields set to {execution_timeout_score}.
- valid_empty: all fields set to {execution_valid_empty_score}, unless the question explicitly expects no rows.
- If there is a material semantic issue, at least one of alignment, logic, or completeness must be <= 4.
- Wrong requested output, wrong metric, wrong filter, wrong join, wrong aggregation, wrong grouping level, wrong ordering direction, or wrong result count should cap affected dimensions at <= 4.
- Extra selected columns should cap completeness at <= 4 unless they are explicitly requested.
- If the candidate contradicts the Hint on a formula, filter, entity, metric, or granularity, cap affected dimensions at <= 4.
>>>
## Score
Use an integer score from -1 to 10.
- 10: correct.
- 8-9: likely correct with only harmless style or alias issues.
- 5-7: partially correct but uncertain or incomplete.
- 1-4: materially wrong.
- 0: invalid or unusable.
- -1: timeout or unavailable execution.

Caps:
- error: score = {execution_error_score}.
- timeout: score = {execution_timeout_score}.
- valid_empty: score = {execution_valid_empty_score}, unless the question explicitly expects no rows.
- Wrong requested output, wrong metric, wrong filter, wrong join, wrong aggregation, wrong grouping level, wrong ordering direction, or wrong result count should score <= 4.
- Extra selected columns should score <= 4 unless explicitly requested.
- If the candidate contradicts the Hint on a formula, filter, entity, metric, or granularity, score <= 4.

...

## Output
Return exactly one valid JSON object and nothing else:
{{
  "cumulative_summary": "<compact updated summary>",
  "evaluations": [
    {{
      "candidate_index": 1,
      "execution_assessment": "valid_nonempty | valid_empty | error | timeout",
\textbf{<<<}
      "scores": {{"alignment": <-1..10>, "schema": <-1..10>, "logic": <-1..10>, "completeness": <-1..10>}},
>>>
      "score": <-1..10>,
      "issues": "<problem plus concrete fix, or 'none' if correct>"
    }}
  ]
}}
\end{lstlisting}

\end{promptbox}

\paragraph{No-group four-score mode (sensitivity study used in \framework).}
This variant is the individual-candidate counterpart of the four-score critic. It removes execution-equivalent grouping and cross-round cumulative-memory fields, while keeping the same alignment/schema/logic/completeness scoring dimensions. Since the judging rubric is otherwise shared with the four-score backbone, we report only the changed interface below.

\begin{promptbox}{Critic Prompt (No-group Four-score Variant)}

\begin{lstlisting}
This no-group variant is identical to the four-score backbone except for the fields shown below.
Unchanged instructions are abbreviated as `...`.
Use `<<<` for the four-score backbone text and `>>>` for the no-group replacement.

<<<
Evaluate a batch of SQL candidates as a strict SQLite Text-to-SQL judge.

...

## Prior cumulative summary
{prior_cumulative_summary}

Use the prior summary only as compact context. Current SQL, question, hint, schema, and execution evidence are authoritative.

## Candidate SQL and Precomputed Execution Evidence
{mode_note}
{candidates_block}
>>>
Evaluate this SQL query as a strict SQLite Text-to-SQL judge.

...

SQL to evaluate:
```sql
{sql}
```

Precomputed execution evidence:
{exec_status}

...

<<<
For each candidate or cluster representative:
1. classify execution,
2. score four dimensions,
3. write one concrete issue summary.

No reference answer is available.
The SQL has already been executed; do not invent execution results.
Treat candidates independently unless their SQL and execution behavior are equivalent.
When the candidate is wrong, `issues` must include both the problem and a repair hint.
>>>
Score the candidate on four dimensions and write one issue summary.
No reference answer is available.
Use the schema, question, hint, SQL, and precomputed execution evidence only.
The SQL has already been executed; do not invent execution results.
...

<<<
## Cumulative summary
Return `cumulative_summary`.
Keep only stable, task-relevant information:
- execution pattern,
- recurring issue pattern,
- repair guidance.
Do not include SQL code.
Do not copy long prior summaries.

## Output
Return exactly one valid JSON object and nothing else:
{{
  "cumulative_summary": "<compact updated summary>",
  "evaluations": [
    {{
      "candidate_index": 1,
      "execution_assessment": "valid_nonempty | valid_empty | error | timeout",
      "scores": {{"alignment": <-1..10>, "schema": <-1..10>, "logic": <-1..10>, "completeness": <-1..10>}},
      "issues": "<problem plus concrete fix, or 'none' if correct>"
    }}
  ]
}}
>>>
## Output
Return exactly one valid JSON object and nothing else:
{{
  "scores": {{"alignment": <-1..10>, "schema": <-1..10>, "logic": <-1..10>, "completeness": <-1..10>}},
  "issues": "<'none' if clearly correct; otherwise problem/risk plus concrete fix>"
}}
\end{lstlisting}

\end{promptbox}

\subsection{\texorpdfstring{Best-of-$K$ Selection}{Best-of-K Selection}}
The Best-of-$K$ module is an optional final-selection add-on rather than a required part of every rollout. It performs listwise comparison over the current top-$K$ candidates and returns a single preferred answer. The design principle is to decouple final selection from upstream utility rank: Candidate 1 may be wrong, execution success alone is insufficient, and the selector must compare output shape, row granularity, filters, joins, aggregation, ordering, limits, and Hint faithfulness. When candidates are semantically and execution-equivalent, the add-on chooses the earlier candidate for determinism. This module is useful for ablations because it tests whether a strict listwise critic can recover from utility ties or mis-scored candidates.

\begin{promptbox}{Best-of-$K$ Selection}

\begin{lstlisting}
You are a strict SQLite Text-to-SQL auditor.

Schema:
{schema}

Question:
{question}

Hint:
{hint}

## Candidate SQL Queries
The candidates are sorted by upstream utility. Candidate 1 may still be wrong.

{candidates_block}

## Task
Choose the single candidate that best answers the question under SQLite execution semantics.

Judge by:
- requested output columns and order
- row granularity and result cardinality
- required entities, filters, and literal values
- joins, grouping, aggregation, arithmetic, ordering, and limits
- execution evidence: errors, timeouts, empty results, wrong shape, NULLs, duplicates, implausible values
- faithfulness to the Hint when it specifies formula, filter, entity, metric, or granularity

Do not choose based on style, aliases, formatting, or upstream rank.
If candidates are semantically and execution-equivalent, choose the earlier candidate.
Execution success alone does not imply correctness.

## Output
Candidate indices are 1-based and must refer only to displayed candidates.
Return exactly one minified JSON object on a single line and nothing else:
{{
  "best_candidate": <single candidate index>,
  "candidate_1_fatal_error": <true or false>,
  "reasoning": "<brief reason, or 'none' if correct>"
}}
\end{lstlisting}

\end{promptbox}

\subsection{SDPO training}

The SDPO stage uses supervised tasks to align the model with both execution-state prediction and SQL reconstruction. The \textbf{induction} direction maps the task context plus an execution-result sketch back to the SQL query. During teacher construction, deduction may receive the privileged gold execution sketch, while induction may receive the privileged gold SQL; these privileged fields are used only as supervision targets and should not be leaked unless they are the requested output.

\begin{promptbox}{SDPO System Prompt}

\begin{lstlisting}
You are an SDPO training model for Text-to-SQL. Given the task context and an execution-result sketch, produce the SQL query that yields that result.

Grounding rules:
- Use only the Schema, Question, Hint, and the specific input in the user prompt.
- Treat Schema, Question, and Hint as authoritative.
- Do not invent tables, columns, filters, joins, calculations, or output columns.
- Treat the specific input as conditioning evidence, not as a replacement for the Schema, Question, or Hint.
- Do not reveal or restate teacher-only privileged information unless the requested output itself is that target.

Output rules:
- Follow the requested output format exactly.
- Return exactly one valid JSON object and nothing else.
- Do not wrap the JSON in markdown fences.
\end{lstlisting}

\end{promptbox}

\begin{promptbox}{SDPO Induction Prompt}

\begin{lstlisting}
Schema:
{schema}

Question:
{question}

Hint:
{hint}

## Execution Result Sketch
{execution_result}

## Direction
Induction: execution-result sketch -> SQL query.

## Task
Write the SQL query for the task using the Schema, Question, Hint, and execution-result sketch.

The execution-result sketch is conditioning evidence. It may help determine output shape, row cardinality, aggregation, ordering, limits, and key values, but it does not override the Schema, Question, or Hint.

Before writing the SQL, ensure that the query:
- uses only schema-defined tables and columns;
- returns only the requested columns, in the requested order;
- matches the requested row granularity and result cardinality;
- uses filters, joins, aggregation, grouping, ordering, HAVING, and LIMIT only when supported by the task;
- preserves the numeric scale implied by the Schema, Question, Hint, and sketch;
- avoids SELECT *, unnecessary DISTINCT, and unsupported calculations.

## Output Format
Return exactly one valid JSON object:
{
  "sql": "SELECT ..."
}
\end{lstlisting}

\end{promptbox}

\begin{promptbox}{SDPO Teacher-Only Privileged Blocks}

\begin{lstlisting}

## Teacher-Only Privileged Gold SQL
```sql
{gold_sql}
```
\end{lstlisting}

\end{promptbox}

\end{document}